\documentclass[manuscript,screen,nonacm,review=false,timestamp=false]{acmart}

\AtBeginDocument{%
  \providecommand\BibTeX{{%
    \normalfont B\kern-0.5em{\scshape i\kern-0.25em b}\kern-0.8em\TeX}}}

\usepackage{xcolor}
\usepackage{multirow}
\usepackage{wasysym} %not allowed
\usepackage{comment}
\usepackage{array}
\usepackage{subcaption}

\newcommand*\rotvertical{\rotatebox{90}}
\usepackage{bigdelim}  %not allowed
\usepackage{mathtools} 
\usepackage{xspace}

\newcommand{\grayc}{\LEFTcircle}
\newcommand{\dos}{Indiscriminate\xspace}

\newcommand{\yes}{\Circle}
\newcommand{\no}{\CIRCLE}
\newcommand{\ie}{{i.e.}}
\newcommand{\eg}{{e.g.}}

\newcommand{\textitparagraph}[1]{\textit{#1}}

\usepackage{pifont}

\setcopyright{acmcopyright}
\copyrightyear{2021}
\acmYear{2021}
\acmDOI{10.1145/1122445.1122456}

\acmBooktitle{}
\newcommand{\xmark}{\ding{55}}
\newcommand{\dashmark}{\rotvertical{\ding{121}}}

\newcommand{\vct}[1]{\boldsymbol{\ensuremath{#1}}}

\newcommand{\set}[1]{\ensuremath{\mathcal{#1}}}

\newcommand{\argmin}{\operatornamewithlimits{\arg\,\min}}

\newcommand{\fullTrain}{{\ensuremath{\set D^\prime}}\xspace}
\newcommand{\cleanTrain}{\ensuremath{\set D}\xspace}
\newcommand{\poisonTrain}{\ensuremath{\set D_p}\xspace}

\newcommand{\cleanVal}{\ensuremath{\set V}\xspace}
\newcommand{\targetVal}{\ensuremath{\set V_t}\xspace}
\newcommand{\fullTest}{\ensuremath{\set T}\xspace}

\newcommand{\poisonLabel}{\ensuremath{y^\prime}}

\newcommand{\modelSymb}{\ensuremath{\set M}\xspace}
\newcommand{\learningSymb}{\ensuremath{\set W}\xspace}
\newcommand{\lossSymb}{\ensuremath{\ell}\xspace}
\newcommand{\LossSymb}{\ensuremath{L}\xspace}
\newcommand{\RegularizedLossSymb}{\ensuremath{\set L}\xspace}

\newcolumntype{P}[1]{>{\centering\arraybackslash}m{#1}}

\newcommand{\psmall}{\includegraphics[width=0.016\textwidth]{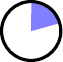}}
\newcommand{\pmedium}{\includegraphics[width=0.016\textwidth]{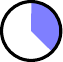}}
\newcommand{\pfull}{\includegraphics[width=0.016\textwidth]{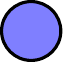}}

\begin{document}


%\title[Wild Patterns Reloaded]{Wild Patterns Reloaded: A Survey of Attacks, Defenses, and Research Challenges for Machine Learning Security against Training Data Poisoning}

\title[Wild Patterns Reloaded]{Wild Patterns Reloaded: A Survey of Machine Learning Security against Training Data Poisoning}

\author{Antonio Emanuele Cinà}
\authornote{Equal contribution}
\email{antonioemanuele.cina@unive.it}
\affiliation{%
  \institution{DAIS, Ca' Foscari University of Venice}
  \streetaddress{Via Torino, 155}
  \city{Venice}
  \country{Italy}
  \postcode{30170}
}

\author{Kathrin Grosse}
\authornotemark[1]
\email{kathrin.grosse@unica.it}
\orcid{0002-5401-4171}
\affiliation{%
  \institution{VITA Lab, \'{E}cole Polytechnique F\'{e}d\'{e}rale de Lausanne}
  %\streetaddress{Via Castelfidardo, 40}
  \city{Lausanne}
  \country{Switzerland}
  \postcode{1015}
}

\author{Ambra Demontis}
\authornote{Corresponding Author}
\email{ambra.demontis@unica.it}
\affiliation{%
  \institution{DIEE, University of Cagliari}
%  \streetaddress{Via Castelfidardo, 1}
%  \city{Cagliari}
  \country{Italy}
%  \postcode{09124}
}

\author{Sebastiano Vascon}
\email{sebastiano.vascon@unive.it}
\affiliation{%
  \institution{DAIS, Ca' Foscari University of Venice and European Center for Living Technology}
%  \streetaddress{Via Torino, 155}
%  \city{Venice}
  \country{Italy}
%  \postcode{30170}
}

\author{Werner Zellinger}
\email{werner.zellinger@scch.at}
\affiliation{%
  \institution{Software Competence Center Hagenberg GmbH (SCCH)}
  \streetaddress{Softwarepark, 21}
  \city{Hagenberg}
  \country{Austria}
  \postcode{4232}
}

\author{Bernhard A. Moser}
\email{bernhard.moser@scch.at}
\affiliation{%
  \institution{Software Competence Center Hagenberg GmbH (SCCH)}
%  \streetaddress{Softwarepark, 21}
%  \city{Hagenberg}
  \country{Austria}
%  \postcode{4232}
}

\author{Alina Oprea}
\email{a.oprea@northeastern.edu}
\affiliation{%
  \institution{Khoury College of Computer Sciences, Northeastern University}
  \streetaddress{360 Huntington Ave}
  \city{Boston}
  \country{MA, USA}
  \postcode{09123}
}

\author{Battista Biggio}
\email{battista.biggio@unica.it}
\affiliation{%
  \institution{DIEE, University of Cagliari, CINI, and Pluribus One}
%  \streetaddress{Piazza d'Armi}
%  \city{Cagliari}
  \country{Italy}
%  \postcode{09123}
}

\author{Marcello Pelillo}
\email{pelillo@unive.it}
\affiliation{%
  \institution{DAIS, Ca' Foscari University of Venice}
%  \streetaddress{Via Torino, 155}
%  \city{Venice}
  \country{Italy}
%  \postcode{30170}
}

\author{Fabio Roli}
\email{fabio.roli@unige.it}
\affiliation{%
  \institution{DIBRIS, University of Genoa, CINI, and Pluribus One}
  \streetaddress{Via All'Opera Pia, 13}
  \city{Genoa}
  \country{Italy}
  \postcode{09124}
}

\renewcommand{\shortauthors}{Cinà, Grosse, et al.}

\begin{abstract}
The success of machine learning is fueled by the increasing availability of computing power and large training datasets.
The training data is used to learn new models or update existing ones, assuming that it is sufficiently representative of the data that will be encountered at test time.
This assumption is challenged by the threat of poisoning, an attack that manipulates the training data to compromise the model's performance at test time.
Although poisoning has been acknowledged as a relevant threat in industry applications, and a variety of different attacks and defenses have been proposed so far, a complete systematization and critical review of the field is still missing. In this survey, we provide a comprehensive systematization of poisoning attacks and defenses in machine learning, reviewing more than 100 papers published in the field in the last 15 years.
We start by categorizing the current threat models and attacks, and then organize existing defenses accordingly. 
While we focus mostly on computer-vision applications, we argue that our systematization also encompasses state-of-the-art attacks and defenses for other data modalities. 
Finally, we discuss existing resources for research in poisoning, and shed light on the current limitations and open research questions in this research field.
\end{abstract}



\maketitle

\section{Introduction}
The unprecedented success of machine learning (ML) in many diverse applications has been inherently dependent on the increasing availability of computing power and large training datasets, under the implicit assumption that such datasets are well representative of the data that will be encountered at test time.
However, this assumption may be violated in the presence of \textit{data poisoning} attacks, i.e., if attackers can either compromise the training data, or gain some control over the learning process (e.g., when model training is outsourced to an untrusted third-party service)~\cite{gu_badnets_2017,perdisciWormSignature2006,Cretu2008Casting-out-Dem,nelson08spam}. Poisoning attacks are staged at training time, and consist of manipulating the training data to degrade the model's performance at test time. 
Three main categories of data poisoning attacks have been investigated so far~\cite{cina2022MLSecurity}. These include indiscriminate, targeted, and backdoor poisoning attacks. In \textit{indiscriminate} poisoning attacks, the attacker manipulates a fraction of the training data to maximize the classification error of the model on the (clean) test samples.
In \textit{targeted} poisoning attacks, the attacker manipulates again a subset of the training data, but this time to cause misclassification of a specific set of (clean) test samples.
In \textit{backdoor} poisoning attacks, the training data is  manipulated by adding poisoning samples containing a specific pattern, referred to as the backdoor trigger, and labeled with an attacker-chosen class label. This typically induces the model to learn a strong correlation between the backdoor trigger and the attacker-chosen class label. Accordingly, at test time, the input samples that embed the trigger are misclassified as samples of the attacker-chosen class.

Although many different attacks can be staged against ML models, a recent survey shows that poisoning is the largest concern for ML deployment in industry~\cite{kumar2020adversarial,grosse2022so}.
Furthermore, several sources confirm that poisoning is already carried out in practice~\cite{grosse2022so,mcgregor2020preventing}.
For example, Microsoft's chatbot Tay\footnote{\url{https://www.theguardian.com/technology/2016/mar/26/microsoft-deeply-sorry-for-offensive-tweets-by-ai-chatbot}} was designed to learn language by interacting with users, but instead learned offensive statements. 
Chatbots in other languages have shared its fate, including a 
 Chinese\footnote{\url{https://www.khaleejtimes.com/technology/ai-getting-out-of-hand-chinese-chatbots-re-educated-after-rogue-rants}} 
and a Korean\footnote{\url{https://www.vice.com/en/article/akd4g5/ai-chatbot-shut-down-after-learning-to-talk-like-a-racist-asshole}} version. Another attack showed how to poison the auto-complete feature in search engines.\footnote{\url{http://www.nickdiakopoulos.com/2013/08/06/algorithmic-defamation-the-case-of-the-shameless-autocomplete/}} 
Finally, a group of extremists submitted wrongly-labeled images of portable ovens with wheels tagging them as \emph{Jewish baby strollers} to poison Google's image search.\footnote{\url{https://www.timebulletin.com/jewish-baby-stroller-image-algorithm/}}
Due to their practical relevance, various scientific articles have been published on training-time attacks against ML models. While the vast majority of the poisoning literature focuses on supervised classification models in the computer vision domain, we would like to remark here that data poisoning has been investigated earlier in cybersecurity~\cite{perdisciWormSignature2006,nelson08spam}, and more recently also in other application domains, like audio~\cite{venomave_aghakhani_2020, Koffas2021Ultrasonic} and natural language processing~\cite{badnl_chen_2020, trojaning_nlp_zhang_2020}, and against different learning methods, such as federated learning~\cite{Bagdasaryan20BackdoorFeder,Xie20DBA}, unsupervised learning~\cite{cina2022clustering,Biggio2013IsDC}, and reinforcement learning~\cite{behzadan2017vulnerability, zhang2020adaptive}.

Within this survey paper, we provide a comprehensive framework for threat modeling of poisoning attacks and categorization of defenses. We identify the main practical scenarios that enable staging such attacks on ML models, and use our framework to properly categorize attacks and defenses. We then review their historical development, also highlighting the main current limitations and the corresponding future challenges. We do believe that our work can serve as a guideline to better understand how and when these attacks can be staged, and how we can defend effectively against them, while also giving a perspective on the future development of trustworthy ML models limiting the impact of malicious users. 
With respect to existing surveys in the literature on ML security, which either consider a high-level overview of the whole spectrum of attacks on ML~\cite{biggio2018wild,chakraborty2018adversarial}, or are specific to an application domain ~\cite{xiao2019characterizing,sun2018adversarial}, our work focuses solely on poisoning attacks and defenses, providing a greater level of detail and a more specific taxonomy.
Other survey papers on poisoning attacks do only consider backdoor attacks~\cite{yansong_backdoor_survey_2020,backdoor_survey_2018,kaviani2021defense}, except for the work by \citet{goldblum2020data} and \citet{Tian2022ACS}. Our survey is complementary to recent work in~\cite{goldblum2020data,Tian2022ACS}; in particular, while in~\cite{goldblum2020data,Tian2022ACS} the authors give an overview of poisoning attacks and countermeasures in centralized and federated learning settings, our survey: (i) categorizes poisoning attacks and defenses in the centralized learning setting, based on a more systematic threat modeling; (ii) introduces a unified optimization framework for poisoning attacks, matches the defenses with the corresponding attacks they prevent, and (iii) discusses the historical timeline of poisoning attacks since the early developments in cybersecurity applications of ML, dating back to more than 15 years ago.

We start our review in Sect.~\ref{sec:threatmodel}, with a detailed discussion on threat modeling for poisoning attacks, and on the underlying assumptions needed to defend against them. This includes defining the learning settings where data poisoning attacks (and defenses) are possible. 
We further highlight the different attack strategies that give us a scaffold for a detailed overview of data poisoning attacks in Sect.~\ref{sec::attacks}. Subsequently, in Sect.~\ref{sec::defenses}, we give an overview of the main defense mechanisms proposed to date against poisoning attacks, including training-time and test-time defense strategies. 
While our survey is mostly focused on poisoning classification models for computer vision, which encompasses most of the work related to poisoning attacks and defenses, in Sect.~\ref{sec:other_domains} 
%Appendix~A 
we discuss related work that has been developed in different contexts.
In Sect.~\ref{sec:resources}, we discuss poisoning research resources such as libraries and dataset containing poisoned models.
Finally, in Sect.~\ref{sec:development} we review the historic development of poisoning attacks and defenses. This overview serves as a basis for discussing ongoing challenges in the field, such as limitations of current threat models, 
the design of more scalable attacks, and the arms race towards designing more comprehensive and effective defenses. For each of these points, we discuss open questions and related future work. 

To summarize, this work provides the following contributions:
(i) we propose a unifying framework for threat modeling of poisoning attacks and systematization of defenses; 
(ii) we categorize around $45$ attack approaches in computer vision according to their assumptions and strategies; 
(iii) we provide a unified formalization for optimizing poisoning attacks via bilevel programming;
(iv) we categorize more than $70$ defense approaches in computer vision, defining six distinct families of defenses; 
%(v) we consider more than $50$ poisoning attacks and defenses in other contexts, e.g., in cybersecurity applications of ML;
(v) we take advantage of our 
framework to match specific attacks with appropriate defenses according to their strategies;
(vi) we discuss state-of-the-art libraries and datasets as resources for poisoning research; and
(vii) we show the historical development of poisoning research and derive open questions, pressing issues, and challenges within the field of poisoning research. Finally, we also derive a unified formalization for optimizing poisoning attacks via bilevel programming, and investigate in the supplementary material in which other domains poisoning attacks and defenses have been developed.
%consider in the \textcolor{red}{supplementary material} more than $50$ poisoning attacks and defenses in other contexts, e.g., in cybersecurity applications of ML.}

%\section{Threat Model and Attack Surface}\label{sec:threatmodel}
\section{Modeling Poisoning Attacks and Defenses}\label{sec:threatmodel}
We discuss here how to categorize poisoning attacks against learning-based systems. We start by introducing the notation and symbols used throughout this paper in Table~\ref{tab:glossary}.
In the remainder of this section, we define the learning settings under which  poisoning attacks have been investigated. We then revisit the framework by \citet{munoz-gonzalez_towards_2017} to systematize poisoning attacks according to the attacker's goal, knowledge of the target system, and capability of manipulating the input data. We conclude by characterizing the defender's goal, knowledge, and capability. 
\begin{table}[t]
\centering
  \renewcommand{\arraystretch}{1}
  \setlength\tabcolsep{1pt} 
  \caption{Notation and symbols used in this survey.}\label{tab:glossary}
\begin{tabular}{@{}cl@{\hskip 0.05in}cl@{\hskip 0.05in}cl@{}}
\toprule
\multicolumn{2}{c}{\textbf{Data}} & \multicolumn{2}{c}{\textbf{Model}} & \multicolumn{2}{c}{\textbf{Noise}} \\ \midrule
\cleanTrain & Clean samples in training set & $\vct\theta$ & Model's parameters & $\vct{t}$ & Test data perturbation \\
\poisonTrain & Poisoning samples in training set & $\phi$ & Model's feature extractor & $\vct{\delta}$ & Training data perturbation \\
\fullTrain & Poisoned training set (\fullTrain$=\cleanTrain \cup \poisonTrain$) & $f$ & Model's classifier & $\Delta$ & Set of admissible manipulations for $\vct\delta$ \\ \cmidrule(l){3-6} 
 &  & \multicolumn{2}{c}{\textbf{Training}} & \multicolumn{2}{c}{\textbf{Attack Strategy}} \\ \cmidrule(l){3-6} 
\cleanVal & Clean samples in validation dataset & \modelSymb & Machine learning model & BL & Bilevel \\
\targetVal & Attacker target samples in validation dataset & \learningSymb & Learning algorithm & FC & Feature Collision \\
 &  & \LossSymb & Loss function & T$^P$ & Patch Trigger \\
\fullTest & Test samples & \RegularizedLossSymb & Training loss (regularized) & T$^S$ & Semantical Trigger \\
$p$ & Percentage of poisoned data & \multicolumn{1}{l}{} &  & T$^F$ & Functional Trigger \\ \bottomrule
\end{tabular}
\end{table}

\subsection{Learning Settings}
\label{sec:learning_setting}

\begin{figure}[t]
\centering
\includegraphics[width=0.97\textwidth]{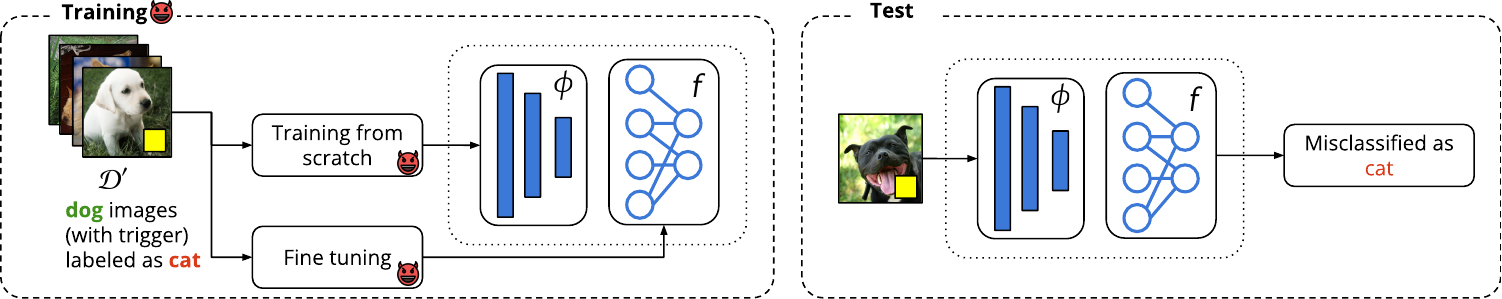}
\caption{Training (left) and test (right) pipeline. The victim collects a training dataset \fullTrain from an untrusted source. The training or fine-tuning algorithm uses these data to train a model \modelSymb, composed of a feature extractor $\phi$, and a classification layer $f$. In the case of fine-tuning, only $f$ is modified, while the feature representation $\phi$ is left untouched.
At test time, some test samples may be manipulated by the attacker to exploit the poisoned model and induce misclassification errors.}
\label{fig:ml_pipeline}
\end{figure}

We define here the three main scenarios under which ML models can be trained, and which can pose serious concerns in relationship to data poisoning attacks. We refer to them below respectively as (i) \emph{training-from-scratch}, (ii) \emph{fine-tuning}, and (iii) \emph{model-training}. In Fig.~\ref{fig:ml_pipeline}, we conceptually represent these settings, along with the entry points of the attack surface which enable staging a poisoning attack. 

\paragraph{Training from Scratch (TS) and Fine Tuning (FT)}
In the \emph{training-from-scratch} and \emph{fine-tuning} scenarios, the user controls the training process, but collects the training data from external repositories, potentially compromised by attackers. In practice, these are the cases where data gathering and labeling represent time-consuming and expensive tasks that not all organizations and individuals can afford, forcing them to collect data from untrusted external sources. The distinction between the two scenarios hinges on how the collected data are employed during training. In the \emph{training-from-scratch} scenario, the collected data is used to train the model from a random initialization of its weights. In the \emph{fine-tuning} setting, instead, a pretrained model is typically downloaded from an untrusted source, and used to map the input samples on a given representation space induced by a feature mapping function $\phi$. Then, a classification function $f$ is fine tuned on top of the given representation $\phi$.

\paragraph{Model Training (MT)}
In the \emph{model-training} (outsourcing) scenario, the user is supposed to have limited computational capacities and outsources the whole training procedure to an untrusted third party, while providing the training dataset. The resulting model can then be provided either as an online service which the user can access via queries, or given directly to the user. In this case, both the feature mapping $\phi$ and the classification function $f$ are trained by the attacker (i.e., the untrusted party). The user, however, can validate the model's accuracy on a separate validation dataset to ensure that the model meets the desired performance requirements.

\subsection{Attack Framework}
\subsubsection{Attacker's Goal}
The goal of a poisoning attack can be defined in terms of the intended security violation, and the attack and error specificity, as detailed below.\smallskip

\textit{Security Violation.} It defines the security violation caused by the attack, which can be: (i) an \textit{integrity} violation, if malicious activities evade detection without compromising normal system operation; (ii) an \textit{availability} violation, if normal system functionality is compromised, causing a denial of service for legitimate users;
or (iii) a \textit{privacy} violation, if the attacker aims to obtain private information about the system itself, its users, or its data.\smallskip

\textit{Attack Specificity.} It determines which samples are subject to the attack. It can be:  (i) \textit{sample-specific} (targeted), if a specific set of sample(s) is targeted by the attack, or (ii) \textit{sample-generic} (indiscriminate), 
if any sample can be affected.\smallskip

\textit{Error Specificity.} It determines how the attack influences the model's predictions. It can be: (i) \textit{error-specific}, if the attacker aims to have a sample misclassified as a specific class; or (ii) \textit{error-generic}, if the attacker attempts to have a sample misclassified as any class different from the true class.\\

\subsubsection{Attacker's Knowledge}~\label{sec:attackerknowledge}
The attacker may get to know some details about the target system, including information about: (i) the (clean) training data \cleanTrain, (ii) the ML model \modelSymb being used, and (iii) the test data \fullTest.
The first component considers how much knowledge the attacker has on the training data.
The second component refers to the ability of the attacker to access the target model, including its internal (trained) parameters $\vct \theta$, but also additional information like hyperparameters, initialization, and the training algorithm.
The third component specifies if the attacker knows in advance (or has access to) the samples that should be misclassified at test time. Although not explicitly mentioned in previous work, we have found that the knowledge of test samples is crucial for some attacks to work as expected. Clearly, attacks that are designed to work on specific test instances are not expected to generalize to different test samples (\eg, to other samples belonging to the same class).
%
%\textcolor{blue}{Why this part is here!?}
%We assume that the model encapsulates a feature extractor, $\phi$, and a classifier, $f$, as shown in Fig.~\ref{fig:ml_pipeline}. The feature extractor transforms the input $\vct x$ into mostly separable numerical features, which are used by the model to output the prediction. For the sake of simplicity, we do not consider the knowledge about the feature set, as in \citet{munoz-gonzalez_towards_2017}, since in this survey we mostly focus on vision applications where color images are commonly represented as three-dimensional tensors. 
%
Depending on the combination of the previously-defined properties, we can define two main attack settings, as detailed below.

\paragraph{White-Box Attacks} The attacker has complete knowledge about the targeted system. Although not always representative of practical cases, this setting allows us to perform a worst-case analysis, and it is particularly helpful for evaluating defenses.

\paragraph{Black-Box Attacks} Black-box attacks can be subdivided into two main categories: black-box \textit{transfer} attacks, and black-box \textit{query} attacks.
Although generally referred to as a black-box attack, \textit{black-box transfer attacks} assume that the attacker has partial knowledge of the training data and/or the target model. In particular, the attacker is assumed to be able to collect a surrogate dataset and use it to train a surrogate model approximating the target. Then, white-box attacks can be computed against the surrogate model, and subsequently \textit{transferred} against the target model. Under some mild conditions, such attacks have been shown to transfer successfully to the target model with high probability~\cite{demontis2019whytransferability}. It is also worth remarking that \textit{black-box query attacks} can also be staged against a target model, by only sending input queries to the model and observing its predictions to iteratively refine the attack, without exploiting any additional knowledge~\cite{tramer16-usenix,papernot17-asiaccs,chen17-aisec}. However, to date, most of the poisoning attacks staged against learning algorithms in black-box settings exploit surrogate models and attack transferability.

\subsubsection{Attacker's Capability}
\label{sec:attackercapability}
The attacker's capability is defined in terms of how the attacker can influence the \textit{learning setting}, and on the \textit{data perturbation} that can be applied to training and/or test samples.

\paragraph{Influence on Learning Setting}~\label{sec:attack_in_settings}
The three learning settings described in Sect.~\ref{sec:learning_setting} open the door towards different data poisoning attacks.
In both \emph{training-from-scratch} (TS) and \emph{fine-tuning} (FT) scenarios, the attacker alters a subset of the training dataset collected and used by the victim to train or fine-tune the machine learning model. 
Conversely, in the \emph{model-training} (MT) scenario, as firstly hypothesized by \citet{gu_badnets_2017}, the attacker acts as a malicious third-party trainer, or as a man-in-the-middle, controlling the training process. The attacker tampers with the training procedure and returns to the victim user a model that behaves according to their goal. The advantage for the attacker is the victim will never be aware of the training dataset actually used. However, to keep their attack stealthy, the attacker must ensure that the provided model retains high prediction accuracy, making sure to pass the validation phase without suspicion from the victim user. 
The attacker's knowledge, discussed in Sect.~\ref{sec:attackerknowledge}, is defined depending on the setting under consideration.
In the \textit{model-training} and \textit{training-from-scratch} settings, \fullTrain and \modelSymb refer to the training data and algorithm used for training the model from random initialization of its weights. Conversely, in the \emph{fine-tuning} setting, \fullTrain and \modelSymb refer to the fine-tuning dataset and learning algorithm, respectively.

\begin{figure}[t]
\centering
\includegraphics[width=1\textwidth]{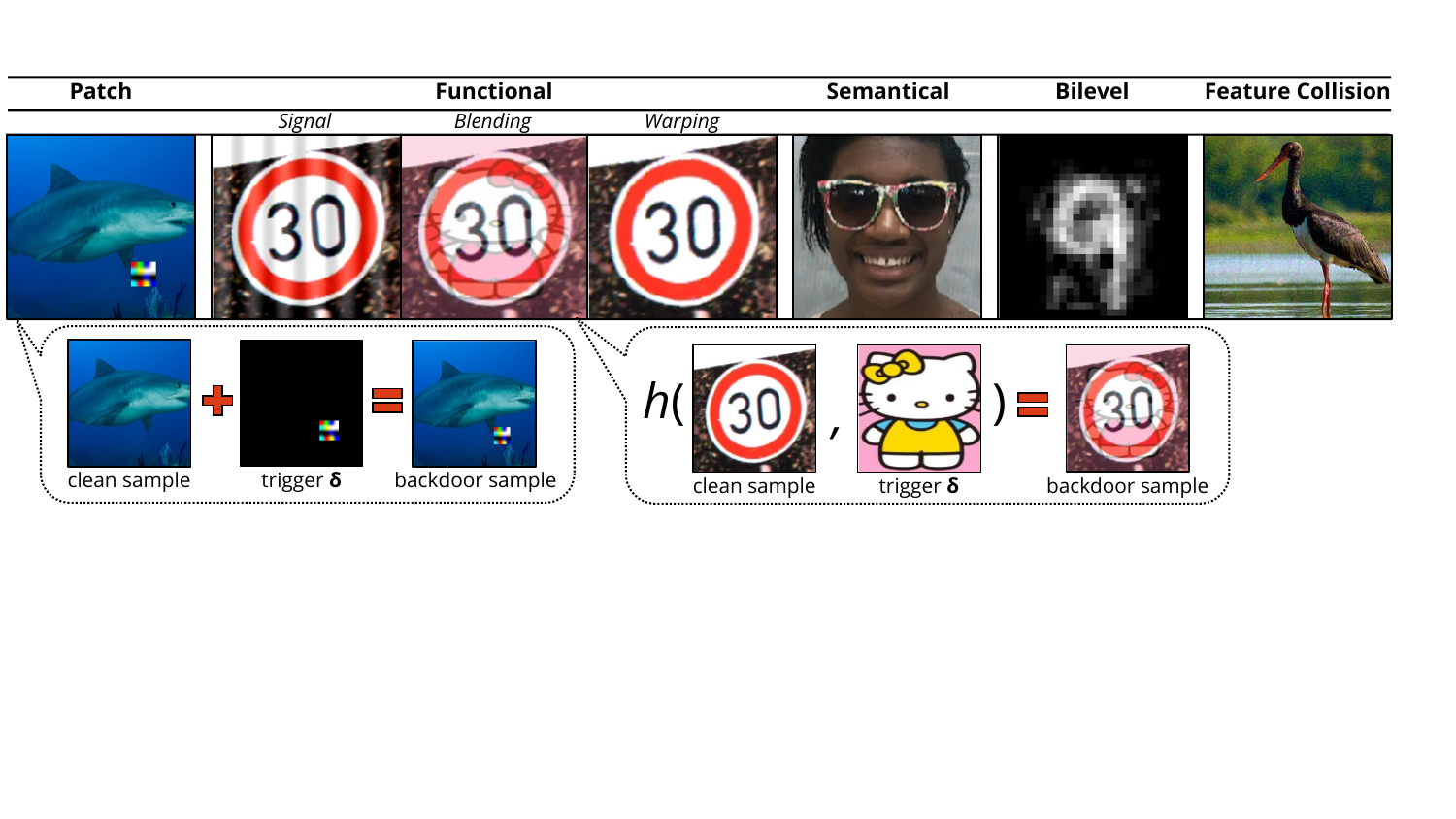}
\caption{%Visual examples of data perturbation noise. The first four figures show some examples of patch, functional, and semantical triggers. For functional triggers we consider signal~\cite{barni_clean_label_2019}, blending~\cite{chen_targeted_2017}, and warping~\cite{WaNet_nguyen_2021} transformations. The remaining two depict poisoning samples crafted with a bilevel attack with visible noise, and a clean-label feature collision attack with imperceptible noise.
Visual examples of data perturbation noise ($\vct \delta$) categories. The first five figures show some examples of patch, functional, and semantical triggers. For functional triggers we consider signal~\cite{barni_clean_label_2019}, blending~\cite{chen_targeted_2017}, and warping~\cite{WaNet_nguyen_2021} transformations. The remaining two depict poisoning samples crafted with a bilevel attack with visible noise, and a clean-label feature collision attack with imperceptible noise. The second row shows the backdoor image generation process with patch and functional blending triggers.
For the latter, a $h$ manipulation function blends the original image and the backdoor trigger according to a certain ratio.}
\label{fig:strategy}
\end{figure}

\paragraph{Data Perturbation} Staging a poisoning attack requires the attacker to manipulate a given fraction ($p$) of the training data. In some cases, i.e., in backdoor attacks, the attacker is also required to manipulate the test samples that are under their control, by adding an appropriate trigger to activate the previously-implanted backdoor at test time.
More specifically, poisoning attacks can alter a fraction of the training labels and/or apply a (different) perturbation to each of the training (poisoning) samples.
If the attack only modifies the training labels, but it does not perturb any training sample, it is often referred to as a \textit{label-flip} poisoning attack.
Conversely, if the training labels are not modified (e.g., if they are validated or assigned by human experts or automated labeling procedures), the attacker can
stage a so-called \textit{clean-label} poisoning attack. Such attacks only slightly modify the poisoning samples, using imperceptible perturbations that preserve the original semantics of the input samples along with their class labels~\cite{shafahi_poison_2018}.
We define the strategies used to manipulate training and test data in poisoning attacks in the next section.

\subsubsection{Attack Strategy}
The attack strategy defines how the attacker manipulates data to stage the desired poisoning attack. Both indiscriminate and targeted poisoning attacks only alter the training data, while backdoor attacks also require embedding the trigger within the test samples to be misclassified. We revise the corresponding data manipulation strategies in the following.
%We revise in the following the data manipulation strategies used to craft poisoning samples.

\paragraph{Training Data Perturbation ($\vct \delta$).}  Two main categories of perturbation have been used to mount poisoning attacks. The former includes perturbations which are found by solving an optimization problem, either formalized as a \textit{bilevel} (BL) programming problem, or as a \textit{feature-collision} (FC) problem. The latter involves the manipulation of training samples in targeted and backdoor poisoning attacks such that they collide with the target samples in the given representation space, to induce misclassification of such target samples in an attacker-chosen class.
When it comes to backdoor attacks, three main types of triggers exist, which can be applied to training samples to implant the backdoor during learning: \textit{patch triggers} (T$^P$), which consist of replacing a small subset of contiguous input features with a patch pattern in the input sample; \textit{functional triggers} (T$^F$), which are embedded into the input sample via a blending function; and \textit{semantical triggers} (T$^S$), which perturb the given input while preserving its semantics (e.g., modifying face images by adding sunglasses, or altering the face expression, but preserving the user identity).
The choice of this strategy plays a fundamental role since it influences the computational effort, effectiveness, and stealthiness of the attack. More concretely, the trigger strategies are less computationally demanding, as they do not require optimizing the perturbation, but the attack may be less effective and easier to detect. Conversely, an optimized approach can enhance the effectiveness and stealthiness of the attack, at the cost of being more computationally demanding. 
In Fig.~\ref{fig:strategy} we give some examples of patch, functional, and semantical triggers, one example of a poisoning attack optimized with bilevel programming, and one example of a \textit{clean-label} feature-collision attack.

\paragraph{Test Data Perturbation ($\vct t$).}
During operation, i.e., at test time, the attacker can submit malicious samples to exploit potential vulnerabilities that were previously implanted during model training, via a backdoor attack. In particular, as we will see in Sect.~\ref{sec:backdoor_poisoning}, backdoor attacks are activated when a specific trigger $\vct t$ is present in the test samples. Normally, the test-time trigger is required to exactly match the trigger implanted during training, thus including patch, functional, and semantical triggers.

%\subsection{Defender's Goal, Knowledge, and Capability}
\subsection{Defense Framework}
\label{sec::defenseSetting}

In this section, we introduce the main strategies that can be used to counter poisoning attacks, based on different assumptions made on the defender's goal, knowledge and capability. 

\subsubsection{Defender's Goal.} The defender's goal is to preserve the integrity, availability, and privacy of their ML model, i.e., to mitigate any kind of security violation that might be caused by an attack. The defender thus adopts appropriate countermeasures to alleviate the effect of possible attacks, without significantly affecting the behavior of the model for legitimate users.

\subsubsection{Defender's Knowledge and Capability.} The defender's knowledge and capability determine in which learning setting a defense can be applied. We identify four aspects that influence how the defender can operate to protect the model: (i) having access to the (poisoned) training data \fullTrain, and to (ii) a separate, clean validation set \cleanVal, and  (iii) having control on the training procedure \learningSymb, and on (iv) the model's parameters $\vct\theta$. We will see in more detail how these assumptions are matched to each defense in Sect.~\ref{sec::defenses}.

\subsubsection{Defense Strategy.} The defense strategy defines how the defender operates to protect the system from malicious attacks before deployment (i.e., at training time),  and after the model's deployment (i.e., at test time). We identify six distinct categories of defenses:
\begin{enumerate}
    \item \textit{training data sanitization}, which aims to remove potentially-harmful training points before training the model;
    \item \textit{robust training}, which alters the training procedure to limit the influence of malicious points;
    \item \textit{model inspection}, which returns for a given model whether it has been compromised (e.g., by a backdoor attack); 
    \item \textit{model sanitization}, which cleans the model to remove potential backdoors or targeted poisoning attempts;  
    \item \textit{trigger reconstruction}, which recovers the trigger embedded in a backdoored network; and
    \item \textit{test data sanitization}, which filters potentially-triggered samples presented at test time.
\end{enumerate}
These defenses essentially work by either (i) cleaning the data or (ii) modifying the model. In the former case, the defender aims to sanitize training or test data. \textit{Training data sanitization} and \textit{test data sanitization} as thus two strategies adopted respectively at training and at test time to mitigate the influence of data poisoning attacks. Alternatively, the defender can act directly on the model, by (i) identifying possible internal vulnerabilities and removing/fixing components that lead to anomalous behavior/classifications, or by (ii) changing the training procedure to make the model less susceptible to training data manipulations. The first approach is employed in \textit{model inspection}, \textit{trigger reconstruction} and \textit{model sanitization} defensive mechanisms. The second approach, instead, includes algorithms that operate at the training level to implement \textit{robust training} mechanisms.  

\subsection{Poisoning Attacks and Defenses}

\begin{figure}[t]
\centering
\includegraphics[width=1\textwidth]{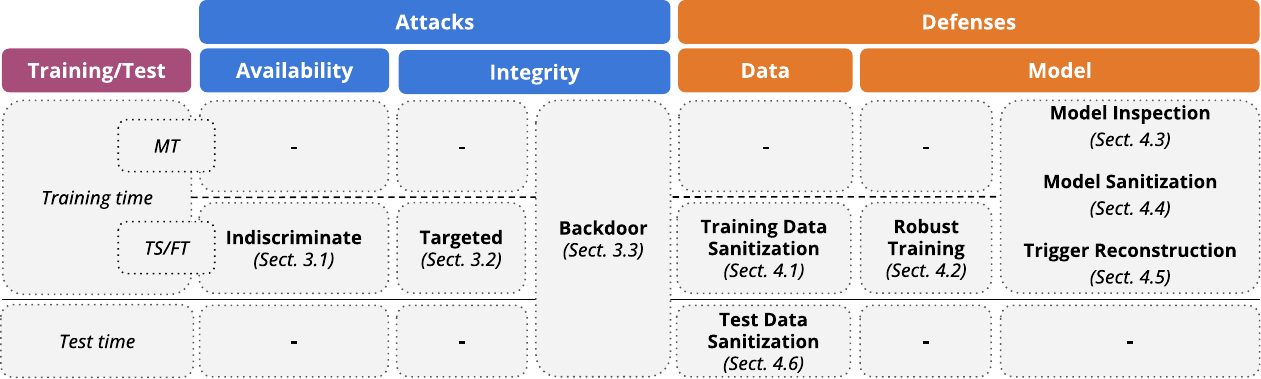}
\caption{Conceptual overview of poisoning attacks and defenses according to our framework. Attacks are categorized based on whether they compromise system integrity or availability. Defenses are categorized based on whether they sanitize data or modify the learning algorithm/model. Training-time (test-time) defenses are applied before (after) model deployment. Training-time interventions are also divided according to whether \emph{model-training} (MT) is outsourced, or \emph{training-from-scratch} (TS) / \emph{fine-tuning} (FT) is performed.}
\label{fig:modeling_attacks_defenses}
\end{figure}

%The attack framework and the defense taxonomy presented so far pave the way towards categorizing poisoning attacks and defenses in a sound manner. 
%Remembering the \emph{first golden rule} seen in \citet{biggio2018wild}, modeling threats against ML systems, and then simulating the corresponding attacks (see the \emph{second golden rule}) is essential to thoroughly evaluate their security against the corresponding attacks. Additionally, it is uttermost important for designing novel defensive techniques which are robust against old attacks and could prevent future ones as explained by the \emph{third golden rule} in \citet{biggio2018wild}, i.e., protect yourself at all times.
We provide in Fig.~\ref{fig:modeling_attacks_defenses} a preliminary, high-level categorization of attacks and defenses according to our framework (while leaving a more complete categorization of each work to Tables~\ref{table:poisoning_categorization}-\ref{table:defensesAll}, respectively for attacks and defenses).
This simplified taxonomy categorizes attacks and defenses based on whether they are applied at training time (and in which learning setting) or at test time; whether the attack aims to violate integrity or availability;\footnote{To our knowledge, no poisoning attack violating a model's privacy has been considered so far, so we omit the privacy dimension from this representation.} and whether the defense aims to sanitize data or modify the learning algorithm/model.
As one may note, indiscriminate and targeted poisoning only manipulate data at training time to violate availability and integrity, respectively, and they are typically staged in the \textit{training-from-scratch} (TS) or \textit{fine-tuning} (FT) learning settings. Backdoor attacks, in addition, require manipulating the test data to embed the trigger and cause the desired misclassifications, with the goal of violating integrity.
Such attacks can be ideally staged in any of the considered learning settings.
For defenses, data sanitization strategies can be applied either at training time or at test time, while defenses that modify the learning algorithm or aim to sanitize the model can be applied clearly only at training time (i.e., before model deployment).
To conclude, while being simplified, we do believe that this conceptual overview of attacks and defenses provides a comprehensive understanding of the main assumptions behind each poisoning attack and defense strategy.
Accordingly, we are now ready to delve into a more detailed description of attacks and defenses in Sects.~\ref{sec::attacks} and~\ref{sec::defenses}.

\section{Attacks}\label{sec::attacks}
We now take advantage of the previous framework to give an overview of the existing attacks according to the corresponding violation and strategy. A compact summary of all attacks from the vision domain is given in Table~\ref{table:poisoning_categorization}.

\begin{figure}[t]
\centering
\begin{subfigure}{0.325\textwidth}
\centering
\includegraphics[width=0.75\textwidth]{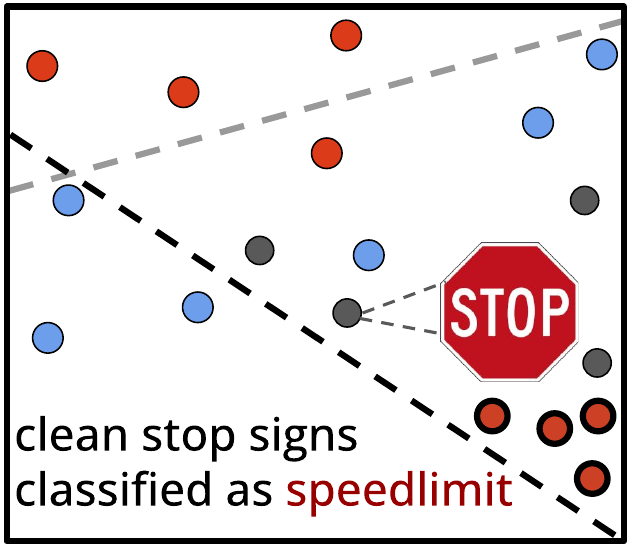}
\caption{Indiscriminate attack.}
\label{fig:dos_conceptual}
\end{subfigure}
\begin{subfigure}{0.325\textwidth}
\centering
\includegraphics[width=0.75\textwidth]{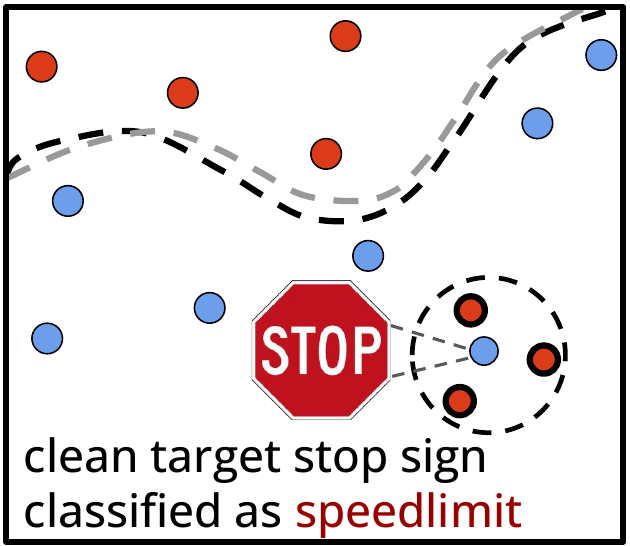}
\caption{Targeted attack.}
\label{fig:targeted_conceptual}
\end{subfigure}
\begin{subfigure}{0.325\textwidth}
\centering
\includegraphics[width=0.75\textwidth]{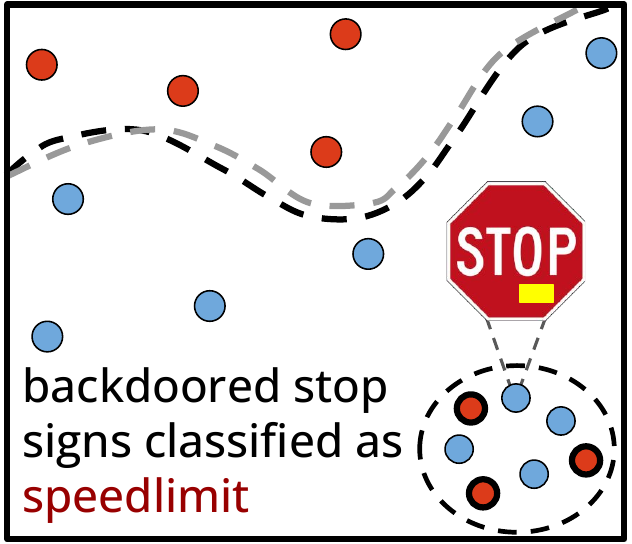}
\caption{Backdoor attack.}
\label{fig:backdoor_conceptual}
\end{subfigure}
\caption{
Conceptual representation of the impact of indiscriminate, targeted, and backdoor poisoning on the learned decision function. We depict the feature representations of the \textit{speed limit} sign (red dots) and \textit{stop signs} (blue dots). The poisoning samples (solid black border) change the original decision boundary (dashed gray) to a poisoned variant (dashed black).
}
\label{fig:poisoning_attacks}
\end{figure}

\begin{table}[t]
 \caption{Taxonomy of existing poisoning attacks, according to the attack framework defined in Sect.~\ref{sec:threatmodel}. The presence of the \checkmark \ indicates that the corresponding properties is satisfied by the attack. For the attacker's knowledge we use: \yes \ when the attacker has knowledge of the corresponding component; \grayc \ if the attacker uses a surrogate to mount the attack; \no \ if the attacker does not require that knowledge. In the attacker's capabilities we use MT, TS and FT as acronyms  for \emph{model-training}, \emph{training-from-scratch}, and \emph{fine-tuning} learning settings. \protect\psmall, \  \protect\pmedium, \ \protect\pfull \  represent the amount of poisoning:  small ($\leq 10\%$), medium ($\leq 30\%$), or high percentage of the training set. 
 The columns $\vct\delta$ and $\vct{t}$ define the training and test strategies: optimized bilevel -- BL,  feature collision -- FC and trigger -- T.}\label{table:poisoning_categorization}
\centering
  \renewcommand{\arraystretch}{0.9}
  \setlength\tabcolsep{1pt} 
\begin{tabular}{lllcccccccc@{\hskip 10pt}ccc}
\toprule
\multirow{2}{*}{}& \multirow{2}{*}{\rotvertical{Section}} & \multicolumn{1}{c}{\multirow{2}{*}{Attacks}} & \multicolumn{2}{c}{Goal} & \multicolumn{3}{c}{Knowledge} & \multicolumn{3}{c}{Capability} &
\multicolumn{2}{c}{Strategy} & Model\\
\cmidrule(r){4-5} \cmidrule(r){6-8}
\cmidrule(r){9-11}\cmidrule(r){12-13} \cmidrule(r){14-14} 

 &  &  & \begin{tabular}[c]{@{}c@{}}Sample\\ Specific\end{tabular} & \begin{tabular}[c]{@{}c@{}}Error\\ Specific\end{tabular} & \cleanTrain & \modelSymb & \fullTest & Setting & \begin{tabular}[c]{@{}c@{}}Clean\\ Label\end{tabular}& p &  $\vct\delta$ & $\vct t$ & DNN\\ 
\midrule
 
\multicolumn{1}{l}{\multirow{10}{*}{\rotvertical{Indiscriminate}}} & \multicolumn{1}{l}{\multirow{2}{*}{\rotvertical{\ref{sec:label_flip}}}} &\citet{biggio11labelnoise}, \citet{labelflip_xiao_2012}, &  &  & \multirow{1}{*}{\yes} & \multirow{1}{*}{\yes} & & \multirow{1}{*}{TS} &  & \pmedium & \multirow{2}{*}{LF} & \multirow{2}{*}{-} &\\ 
& &  \citet{labelcontamination_Xiao_2015}, \citet{paudice2018label} &  &  & \multirow{1}{*}{\yes} & \multirow{1}{*}{\yes}&   & \multirow{1}{*}{TS} &  & \pmedium \\ 
 
 \addlinespace
 
 &  \multicolumn{1}{l}{\multirow{5}{*}{\rotvertical{\ref{sec:dog_grad_flip}}}} & \citet{biggio2012poisoning, battista_secure_features2015} &  &  & \multirow{2}{*}{\yes} & \multirow{2}{*}{\yes} &  & \multirow{2}{*}{TS} & & \multirow{2}{*}{\pmedium} &  \multirow{5}{*}{BL} & \multirow{5}{*}{-} &\\
 & & \citet{frederickson2018attack} &&\\
& & BetaPoison \cite{cina2021hammer}, \citet{private_learners_ma_2019} &  & \checkmark & \yes & \grayc& & TS & & \pmedium & \\
 & & \citet{solans2020poisoning-fairness, demontis2019whytransferability} &  &  & \grayc & \grayc & & TS & & \multicolumn{1}{l}{\pmedium} & \multicolumn{1}{l}{} \\
 & & \citet{munoz-gonzalez_towards_2017, yang_generative_2017} &  &  & \grayc & \grayc & & TS & & \pmedium & && \checkmark \\
 
 \addlinespace

  & \multicolumn{1}{l}{\multirow{3}{*}{\rotvertical{\ref{sec:dos_clean_label}}}} & \citet{using_machine_learning}  &  & \checkmark &\yes & \yes&  & TS & \checkmark & \pfull & \multirow{3}{*}{BL} & \multirow{3}{*}{-} & \\
  & & \citet{confuse_fengCZ_2019} &  & \checkmark & \grayc & \grayc &  & TS & \checkmark & \pfull &  &  &\checkmark\\
 & & \citet{preventing_fowl_2021} &  & \checkmark & \grayc & \grayc & & TS & \checkmark & \pfull &  & &\checkmark\\
 
\addlinespace
\midrule
\addlinespace

\multicolumn{1}{l}{\multirow{8}{*}{\rotvertical{Targeted}}} & \multicolumn{1}{l}{\multirow{3}{*}{\rotvertical{\ref{sec:targeted_optimized}}}} & \citet{koh_understanding_2017} & \checkmark & \checkmark & \yes & \yes &\multirow{1}{*}{\checkmark} & FT & & \psmall&  \multirow{3}{*}{BL} & \multirow{3}{*}{-}&\\
& & \citet{munoz-gonzalez_towards_2017} & \checkmark & \checkmark & \grayc & \grayc & & TS &  & \psmall & & & \checkmark \\
& & \citet{subpopulation_jagielski_2020} &  & \checkmark & \grayc & \grayc & & TS/FT &  & \psmall & & & \checkmark \\

\addlinespace

 & \multicolumn{1}{l}{\multirow{3}{*}{\rotvertical{\ref{sec:targeted_featurecollision}}}}  & PoisonFrog \cite{shafahi_poison_2018} & \checkmark & \checkmark & \yes & \yes& \checkmark & FT & \checkmark & \psmall & \multirow{3}{*}{FC} & \multirow{3}{*}{-}& \checkmark\\
 &  & \citet{guo_practical_2020}, StingRay\cite{suciu2018does} & \multirow{2}{*}{\checkmark} & \multirow{2}{*}{\checkmark} & \multirow{2}{*}{\grayc} & \multirow{2}{*}{\grayc}& \multirow{2}{*}{\checkmark} & \multirow{2}{*}{FT} & \multirow{2}{*}{\checkmark} & \multirow{2}{*}{\psmall} & & & \multirow{2}{*}{\checkmark}\\
 &  & ConvexPolytope~\cite{zhu_transferable_2019}, BullseyePolytope~\cite{aghakhani20bullseye} &  \\
  \addlinespace 

 & \multicolumn{1}{l}{\multirow{2}{*}{\rotvertical{~\ref{sec:targeted_optimized_clean_label}}}} &\multirow{2}{*}{\citet{geiping_witches_2020}, MetaPoison~\cite{huang_metapoison_2020}} & \multirow{2}{*}{\checkmark} & \multirow{2}{*}{\checkmark} & \multirow{2}{*}{\grayc} & \multirow{2}{*}{\grayc} & \multirow{2}{*}{\checkmark} & \multirow{2}{*}{TS} & \multirow{2}{*}{\checkmark} & \multirow{2}{*}{\psmall}  & \multirow{2}{*}{BL} & \multirow{2}{*}{-} & \multirow{2}{*}{\checkmark}\\
 & & & & & & & & & & & & &\\
\midrule
\addlinespace

\multirow{16}{*}{\rotvertical{Backdoor}} & \multicolumn{1}{l}{\multirow{3}{*}{\rotvertical{\ref{sec:backdoor_patch}}}} & BadNet~\cite{gu_badnets_2017}, LatentBackdoor~\cite{yao_latent_2019}&  \checkmark & \checkmark & \yes & \yes & & MT &  & \psmall & \multirow{3}{*}{T$^P$} & \multirow{3}{*}{T$^P$}& \checkmark\\
 & & BaN \cite{dynamic_backdoor_salem_2020} & \checkmark & \checkmark & \yes & \yes& & MT &  & \pmedium & & & \checkmark \\
 & & TrojanNN \cite{liu_trojaning_2018} & \checkmark & \checkmark & \no & \yes & & MT &  & \psmall & & & \checkmark \\

 \addlinespace
 
 & \multicolumn{1}{l}{\multirow{4}{*}{\rotvertical{\ref{sec:backdoor_functional}}}}  & WaNET~\cite{WaNet_nguyen_2021}, \citet{steganography_shaofeng_2020}, DFST~\cite{Cheng2021DFST} & \checkmark & \checkmark & \yes & \yes& & MT &  & \pmedium &\multirow{4}{*}{T$^F$} & \multirow{4}{*}{T$^F$} & \checkmark  \\
 &  &Refool \cite{reflection_backdoor_liu_2020} & \checkmark & \checkmark & \yes & \yes & & TS & \checkmark &\pmedium & & &\checkmark\\
  &  &SIG \cite{barni_clean_label_2019} & \checkmark & \checkmark & \no & \no & & TS & \checkmark & \pmedium &  & & \checkmark\\
    %&  &\citet{invisible_zhong_2020} & \checkmark & \checkmark & \grayc & \grayc &  & TS/FT &  & \psmall &  & &\checkmark\\
  &  &\citet{chen_targeted_2017, invisible_zhong_2020} & \checkmark & \checkmark & \no & \no &  & TS/FT &  & \psmall &  & &\checkmark\\
  
  \addlinespace
  
  & \multicolumn{1}{l}{\multirow{3}{*}{\rotvertical{\ref{sec:backdoor_semantical}}}} &FaceHack \cite{facehack_sarkar_2020} & \checkmark & \checkmark & \yes & \yes & & MT &  & \pmedium & \multirow{3}{*}{T$^S$} & \multirow{3}{*}{T$^S$} &\checkmark\\
 &  &\citet{chen_targeted_2017} & \checkmark & \checkmark & \no & \no &  & TS/FT &  &\psmall &   & &\checkmark\\
 &  &\citet{wenger_backdoor_2020} & \checkmark & \checkmark & \yes & \no &  & FT &  & \pmedium  &  & &\checkmark\\

  \addlinespace
  
 &  \multicolumn{1}{l}{\multirow{4}{*}{\rotvertical{\ref{sec:backdoor_optimized_functional}}}} &\citet{input_aware_nguyen_2020}, LIRA~\cite{Doan_2021_ICCV} & \checkmark & \checkmark & \yes & \yes & & MT &  & \pmedium &  \multirow{4}{*}{BL} & \multirow{4}{*}{T$^F$} &\checkmark\\
  &  &\citet{steganography_shaofeng_2020} & \checkmark & \checkmark & \no & \yes & & MT &  & \psmall &  & &\checkmark\\
   &  &\citet{Li2021Invisible} & \checkmark & \checkmark & \yes & \grayc &  & TS &  & \pmedium &  & &\checkmark\\
 &  &\citet{invisible_zhong_2020} & \checkmark & \checkmark & \grayc & \grayc &  & TS/FT  &  & \psmall &  & &\checkmark\\
  
  \addlinespace 
  
  %&  \multicolumn{1}{l}{\multirow{2}{*}{\rotvertical{~\ref{sec:backdoor_fc_patch}}}}  & 
  %\multirow{2}{*}{HiddenTrigger~\cite{saha_hidden_2019}} & \multirow{2}{*}{\checkmark} & \multirow{2}{*}{\checkmark} & \multirow{2}{*}{\yes} & \multirow{2}{*}{\yes} &  & \multirow{2}{*}{FT} & \multirow{2}{*}{\checkmark} & \multirow{2}{*}{\pmedium} & \multirow{2}{*}{FC} & \multirow{2}{*}{T$^P$}& \multirow{2}{*}{\checkmark}\\
  % & & & & & & & & & & & & &\\
&  \multicolumn{1}{l}{\multirow{2}{*}{\rotvertical{~\ref{sec:backdoor_fc_patch}}}}  & 
  HiddenTrigger~\cite{saha_hidden_2019} & \checkmark & \checkmark & \yes & \yes &  & FT & \checkmark & \pmedium & \multirow{2}{*}{FC} & \multirow{2}{*}{T$^P$}& \checkmark\\
 & &\citet{turner_clean-label} & \checkmark & \checkmark & \grayc & \grayc &  & TS & \checkmark & \psmall &  & & \checkmark\\

 &  \multicolumn{1}{l}{\multirow{2}{*}{\rotvertical{\ref{sec:backdoor_optimized_patch}}}} &\multirow{2}{*}{\citet{sleeper_goldstein_2021}} & \multirow{2}{*}{\checkmark} & \multirow{2}{*}{\checkmark} & \multirow{2}{*}{\grayc} & \multirow{2}{*}{\grayc} &  & \multirow{2}{*}{TS} & \multirow{2}{*}{\checkmark} & \multirow{2}{*}{\psmall} & \multirow{2}{*}{BL} & \multirow{2}{*}{T$^P$} & \multirow{2}{*}{\checkmark}\\
    & & & & & & & & & & & & &\\

\bottomrule
\end{tabular}
\end{table}

\subsection{Indiscriminate (Availability) Poisoning Attacks}
\label{sec:availability_poisoning}
Indiscriminate poisoning attacks represent the first class of poisoning attacks against ML algorithms. 
The attacker aims to subvert the system functionalities, compromising its availability for legitimate users by poisoning the training data. More concretely, the attacker's goal is to cause misclassification on clean validation samples by injecting new malicious samples or perturbing existing ones in the training dataset. In Fig.~\ref{fig:dos_conceptual} we consider the case where an attacker poisons a linear street-sign classifier to have stop signs misclassified as speed limits. 
The adversary injects poisoning samples to rotate the classifier's decision boundary, thus compromising the victim's model performance. 
In the following, we present the strategies developed in existing works, and we categorize them in Table~\ref{table:poisoning_categorization}. 
Although they could also operate on the \emph{fine-tuning} (FT) scenario, existing works have been proposed only in the \emph{training-from-scratch} (TS) setting. By contrast, their application in the \emph{model-training} (MT) scenario would not be feasible, as the model, with reduced accuracy due to the attack, would not pass the user validation phase. Indiscriminate attacks, to be adaptable in the latter scenario, must compromise the availability of the system but not in terms of increasing the classification error. This has been recently done by~\citet{cina2022Sponge}, who proposed a so-called \textit{sponge} poisoning attack aimed to increase the model's prediction latency.

\subsubsection{Label-Flip Poisoning}~\label{sec:label_flip}
The most straightforward strategy to stage poisoning attacks against ML is label-flip, originally proposed by~\citet{biggio11labelnoise}. The adversary does not perturb the feature values, but they mislabel a subset of samples in the training dataset, compromising the performance accuracy of ML models such as Support Vector Machines (SVMs).
Beyond that, \citet{labelflip_xiao_2012} showed that random flips could have far-from-optimal performance, which nevertheless would require solving an NP-hard optimization problem. 
Due to its intractability, heuristic strategies have been proposed by \citet{labelflip_xiao_2012}, and later by  \citet{labelcontamination_Xiao_2015}, to efficiently approximate the optimal formulation.

\subsubsection{Bilevel Poisoning}~\label{sec:dog_grad_flip}
In this case, the attacker manipulates both the training samples and their labels. The pioneering work in this direction was proposed by \citet{biggio2012poisoning}, where a gradient-based indiscriminate poisoning attack is exploited against SVMs. They exploited \textit{implicit differentiation} to derive the gradient required to optimize the poisoning samples by their iterative algorithm. Until convergence, the poisoning samples are iteratively updated following the implicit gradient, directing towards maximization of the model's validation error. Mathematically speaking, this idea corresponds to treating the poisoning task as a bilevel optimization problem:
\begin{comment}
%before rebuttal
\begin{eqnarray}
    \label{eq:dos_outer_problem}
    \max_{\|\vct \delta\|_q \leq \epsilon} &&  \LossSymb(\cleanVal, \modelSymb, \vct\theta^\star) \, ,\\
    \label{eq:dos_inner_problem}
    {\rm s.t.}  && \vct\theta^\star \in \argmin_{\vct \theta} \, 
    \RegularizedLossSymb(\cleanTrain \cup \poisonTrain^{\vct \delta}, \modelSymb,  \vct\theta) \, .
\end{eqnarray}
\end{comment}
\begin{eqnarray}
    \label{eq:dos_outer_problem}
    \max_{\vct \delta \in \Delta} &&  \LossSymb(\cleanVal, \modelSymb, \vct\theta^\star) \, ,\\
    \label{eq:dos_inner_problem}
    {\rm s.t.}  && \vct\theta^\star \in \argmin_{\vct \theta} \, 
    \RegularizedLossSymb(\cleanTrain \cup \poisonTrain^{\vct \delta}, \modelSymb,  \vct\theta) \, .
\end{eqnarray}
with $\Delta$ being the set of admissible manipulation of the training samples that preserve the constraints imposed by the attackers (\eg, $\ell_p$, or box-constraints)\footnote{For example, the attacker can constraint the perturbation magnitude of $\vct \delta$ imposing $\| \vct \delta\|_p \leq \epsilon$ with $\Delta = \{\vct \delta \in \mathbb{R}^{n\times d} ~|~ \|\vct \delta\|_p \leq \epsilon\}$.}. We define with $\poisonTrain = \{(\vct x_i, y_i)\}_{i=1}^{n}$ the training data controlled by the attacker, before any perturbation is applied, being $y_i$ the pristine label of sample $\vct x_i$ and $n$ the number of samples in \poisonTrain. We then denote with $\poisonTrain^{\vct\delta}$ the corresponding poisoning dataset manipulated according to the perturbation parameter $\vct \delta$.
The attacker optimizes the perturbation $\vct{\delta}$ (applied to the poisoning samples \poisonTrain) to increase the error/loss \LossSymb of the target model \modelSymb on the clean validation samples \cleanVal. Our formulation in Eqs.~\eqref{eq:dos_outer_problem}-\eqref{eq:dos_inner_problem} encompass both dirty or clean-label attacks according to the nature of $\poisonTrain^{\vct\delta}$.
%Given $\poisonTrain = \{(\vct x_i, y_i)\}_{i=1}^{n_p}$, with $y_i$ being the pristine label of $\vct x_i$ and $n_p$ the number of samples in \poisonTrain, we define $\poisonTrain^{\vct\delta} = \{(\vct x_i + \vct \delta_i, y_i^\prime)\}_{i=1}^{n_p}$\footnote{\textcolor{red}{In this example we used $\vct \delta$ as additive noise. To be more generic we can define a manipulation function $h$ parametrized by $\vct \delta$ and the sample $\vct x$ to perturb.}}, being $y_i^\prime$ the poisoning label chosen by the attacker. Therefore, for a clean-label we set $y_i^\prime = y_i$ while for dirty-label attacks we set $y_i^\prime \neq y_i$.
For example, we can define $\poisonTrain^{\vct\delta} = \{(\vct x_i + \vct \delta_i, y_i^\prime)\}_{i=1}^{n}$\footnote{In this example we used $\vct \delta$ as additive noise. To be more generic we can define a manipulation function $h$ parametrized by $\vct \delta$ and the sample $\vct x$ to perturb. See example in Fig.~\ref{fig:strategy} for functional blending trigger.}, being $y_i^\prime$ the poisoning label chosen by the attacker, with $y_i^\prime = y_i$ for a clean-label attack and $y_i^\prime \neq y_i$ for a dirty-label attack.
%Given $\poisonTrain^{\vct\delta} = \{(\vct x_i + \vct \delta_i, y_i^p)\}_{(\vct x_i, y_i) \in \poisonTrain}$, with $y_i$ being the pristine label of $\vct x_i$ and $y_i^p$ the poisoning label, for a clean-label we set $y_i^p = y_i$ while for dirty-label attacks we set $y_i^p \neq y_i$. %Furthermore, the attacker can constraint the perturbation magnitude of $\vct \delta$ (\ie, imposing $\| \vct \delta\|_q \leq \epsilon$) to make it less visible and therefore more stealthy against inspection.
Solving this bilevel optimization is challenging, since the inner and the outer problems in Eqs.~\eqref{eq:dos_outer_problem}-\eqref{eq:dos_inner_problem} have conflicting objectives. More concretely, the inner objective is a regularized empirical risk minimization, while the outer one is empirical risk maximization, both considering data from the same distribution. 
A similar approach was later generalized in \citet{battista_secure_features2015} and \citet{frederickson2018attack} to target feature selection algorithms (\ie, LASSO, ridge regression, and elastic net). 
Subsequent work tried to analyze the robustness of ML models when the attacker has limited knowledge about the training dataset or the victim's classifier.
In this scenario, the most investigated methodology is given by the \textit{transferability} of the attack~\cite{demontis2019whytransferability, solans2020poisoning-fairness, private_learners_ma_2019}. The attacker crafts the poisoning samples using surrogate datasets and/or models, and then transfers the attack to another target model. This approach has proven effective for corrupting logistic classifiers \cite{demontis2019whytransferability}, algorithmic fairness \cite{solans2020poisoning-fairness}, and differentially-private learners \cite{private_learners_ma_2019}. More details about the transferability of poisoning attacks are reported in Sect.~\ref{sec:transferability}.

Differently from previous work, \citet{cina2021hammer} observed that a simple heuristic strategy, together with a variable reduction technique, can reach noticeable results against linear classifiers, with increased computational efficiency. %than bilevel attacks.
More concretely, the authors showed how previous gradient-based approaches can be affected by several factors (\eg, loss landscape) that degrade their performance in terms of computation time and attack efficiency.

Although effective, the aforementioned poisoning attacks have been designed to fool models with a relatively small number of parameters. More recently, \citet{munoz-gonzalez_towards_2017} showed that devising poisoning attacks against larger models, such as convolutional neural networks, can be computationally and memory demanding. 
To this end, \citet{munoz-gonzalez_towards_2017} pioneered the idea to adapt hyperparameter optimization methods, which aims to solve bilevel programming problems more efficiently, in the context of poisoning attacks. The authors indeed proposed a back-gradient descent technique to optimize poisoning samples, drastically reducing the attack complexity.
The underlying idea is to back-propagate the gradient of the objective function to the poisoning samples while learning the poisoned model. However, they assume the objective function is sufficiently smooth to trace the gradient backward correctly.
%Accordingly with the results in \cite{munoz-gonzalez_towards_2017}, \citet{yang_generative_2017} showed that computing the analytical gradient of the validation loss in Eq.~\eqref{eq:dos_outer_problem} with respect to the poisoning samples requires computing expensive operations. Nevertheless, they showed that even estimating it can be as well computational and query expensive. 
Accordingly with the results in \cite{munoz-gonzalez_towards_2017}, \citet{yang_generative_2017} showed that computing the analytical or estimated gradient of the validation loss in Eq.~\eqref{eq:dos_outer_problem} with respect to the poisoning samples can be as well computational and query expensive. 
Another way explored in \citet{yang_generative_2017} was to train a generative model from which the poisoning samples are generated, thus increasing the generation rate.

\subsubsection{Bilevel Poisoning (Clean-Label)}~\label{sec:dos_clean_label}
Previous work examined in Sect.~\ref{sec:dog_grad_flip} assumes that the attacker has access to a small percentage of the training data and can alter both features and labels. Similar attacks have been staged by assuming that the attacker can control a much larger fraction of the training set, while only slightly manipulating each poisoning sample to preserve its class label, i.e., performing a clean-label attack.
This idea was introduced by~\citet{using_machine_learning}, who considered manipulating the whole training set to arbitrarily define the importance of individual features on the predictions of convex learners. 
More recently, DeepConfuse~\cite{confuse_fengCZ_2019} and \citet{preventing_fowl_2021} proposed novel techniques to mount clean-label poisoning attacks against DNNs. In \cite{confuse_fengCZ_2019}, the attacker trains a generative model, similarly to~\cite{yang_generative_2017}, to craft clean-label poisoning samples which can compromise the victim's model. 
Inspired by recent developments proposed in \cite{geiping_witches_2020}, \citet{preventing_fowl_2021} used a gradient alignment optimization technique to alter the training data imperceptibly, but diminishing the model's performance.
Even though \citet{confuse_fengCZ_2019} and \citet{preventing_fowl_2021} can target DNNs, the attacker is assumed to perturb a high fraction of samples in the training set. 
We do believe that this is a very demanding setting for poisoning attacks. In fact, such attacks are often possible because ML is trained on data collected in the wild (\eg, labeled through tools such as a mechanical Turk) or crowdsourced from multiple users; thus, it would be challenging for attackers in many applications to realistically control a substantial fraction of these training data. 
In conclusion, the quest for scalable, effective, and practical indiscriminate poisoning attacks on DNNs is still open. Accordingly, it remains also unclear whether DNNs can be significantly subverted by such attacks in practical settings.

\subsection{Targeted (Integrity) Poisoning Attacks}
\label{sec:targeted_poisoning}
In contrast to indiscriminate poisoning, targeted poisoning attacks preserve the availability, functionality and behavior of the system for legitimate users, while causing misclassification of some specific target samples. Similarly to indiscriminate poisoning, targeted poisoning attacks manipulate the training data but they do not require modifying the test data. 

An example of a targeted attack is given in  Fig.~\ref{fig:targeted_conceptual}, where the classifier's  decision function for clean samples is not significantly changed after poisoning, preserving the model's accuracy. However, the model isolated the target stop sign (grey) to be misclassified as a speed-limit sign. The system can still correctly classify the majority of clean samples, but outputs wrong predictions for the target stop sign.

In the following sections, we describe the targeted poisoning attacks categorized in Table~\ref{table:poisoning_categorization}. Notably, such attacks have been investigated both in the \emph{training-from-scratch} (TS) and \emph{fine-tuning} (FT) settings, defined in Sect.~\ref{sec:learning_setting}.

\subsubsection{Bilevel Poisoning}~\label{sec:targeted_optimized}
In Sect.~\ref{sec:dog_grad_flip}, we reviewed the work in \citet{munoz-gonzalez_towards_2017}. In addition to indiscriminate poisoning, the authors also formulated targeted poisoning attacks as: 
\begin{eqnarray}
    \label{eq:targeted_outer_problem}
    %\min_{\vct \delta} &&  \LossSymb(\cleanVal, \cup \targetVal, \vct\theta^\star) \, ,\\
    \min_{\vct \delta \in \Delta} &&  \LossSymb(\cleanVal, \modelSymb, \vct\theta^\star) + \LossSymb(\targetVal, \modelSymb,\vct\theta^\star) \, ,\\
    \label{eq:targeted_inner_problem}
    \text{s.t.} && \vct\theta^\star \in  \argmin_{\vct \theta} \,  \RegularizedLossSymb(\cleanTrain \cup \poisonTrain^{\vct \delta},\modelSymb, \vct\theta) \,  .
\end{eqnarray}
Within this formulation, the attacker optimizes the  perturbation $\vct{\delta}$ on the poisoning samples \poisonTrain to have a set of target (validation) samples \targetVal misclassified, while preserving the accuracy on the clean (validation) samples in \cleanVal.
It is worth remarking here that the attack is optimized on a set of validation samples, and then evaluated on a separate set of test samples. The underlying rationale is that the attacker can not typically control the specific realization of the target instances at test time (e.g., if images are acquired from a camera sensor, the environmental and acquisition conditions can not be controlled), and the attack is thus expected to generalize correctly to that case. 

A similar attack was introduced by \citet{koh_understanding_2017}, to show the equivalence between gradient-based (bilevel) poisoning attacks and influence functions, i.e., functions defined in the area of robust statistics that identify the most relevant training points influencing specific predictions. Notably, these authors were the first to consider the \emph{fine-tuning} (FT) scenario in their experiments, training the classification function $f$ (i.e., an SVM with the RBF kernel) on top of a feature representation $\phi$ extracted from an internal layer of a DNN. 
Although these two bilevel optimization strategies have been proven effective, they remain too computationally demanding to be applied to DNNs.

\citet{subpopulation_jagielski_2020} showed how to generalize targeted poisoning attacks to an entire subpopulation in the data distribution, while reducing the computational cost. To create subpopulations, the attacker selects data samples by matching their features or clustering them in feature space. The poisoning attack can be performed either by label flipping, or linearizing the influence function to approximate the poisoning gradients, thus reducing the computational cost of the attack.
\citet{munoz-gonzalez_towards_2017} and \citet{subpopulation_jagielski_2020} define a more ambitious goal for the attack compared to \citet{koh_understanding_2017}, as their attacks aim to generalize to all samples coming from the target distribution or the given subpopulation. Specifically, the attack by \citet{koh_understanding_2017} is tailored for misleading the model only for some specific test samples, which means considering the test set \fullTest rather than a validation set \targetVal in Eq.~\eqref{eq:targeted_outer_problem}.
However, the cost of the attack by~\citet{munoz-gonzalez_towards_2017} is quite high, due to need of solving a bilevel problem, while the attack by \citet{subpopulation_jagielski_2020} is faster, but it does not achieve the same success rate on all subpopulations.

\subsubsection{Feature Collision (Clean-Label)}~\label{sec:targeted_featurecollision}
This category of attacks is based on a heuristic strategy named \textit{feature collision}, suited to the so-called \emph{fine-tuning} (FT) scenario, which avoids the need of solving a complex bilevel problem to optimize poisoning attacks. In particular, PoisonFrog~\cite{shafahi_poison_2018} was the first work proposing this idea, which can be formalized as:%\footnote{We neglect the penalty term used to increase the attack's stealthiness introduced in \cite{shafahi_poison_2018} as it is not related to the \textit{feature collision} strategy.}
\begin{comment}
\begin{eqnarray}
    \label{eq:feature_collision}
    \min_{\|\vct \delta\|_p \leq \epsilon} &  \|\phi(\vct x +\vct \delta) - \phi(\vct z)\|_2^2 \,.
\end{eqnarray}
\end{comment}
\begin{eqnarray}
    \label{eq:feature_collision}
    \min_{\vct \delta \in \Delta} &  \|\phi(\vct x +\vct \delta) - \phi(\vct z)\|_2^2 \,.
\end{eqnarray}
This attack amounts to creating a poisoning sample $\vct x + \vct \delta$ that collides with the target test sample $\vct z \in \fullTest$ in the feature space, so that the fine-tuned model predicts $\vct z$ according to the poisoning label associated with $\vct x$. To this end, the adversary leverages the feature extractor $\phi$ to minimize the distance of the poisoning sample with the target in the feature space. 
%Moreover, the authors observed that, due to the complexity and nonlinear behavior of $\phi$, the poisoning samples only need to be slightly perturbed, and that even samples coming from different distributions can be only slightly perturbed to match the feature representation of the target sample. \textcolor{red}{Therefore, the adversarial noise $\vct \delta$ is bounded by the attacker in $\ell_q$ norm to encompass clean-label attack and stealthiness against human inspection. \citet{shafahi_poison_2018} incorporate such }
Moreover, the authors observed that, due to the complexity and nonlinear behavior of $\phi$, even poisoning samples coming from different distributions can be slightly perturbed in the input space to match the feature representation of the target sample $\vct z$. To make the poisoning sample
look realistic in input space and implement a clean-label attack, the adversarial perturbation $\vct \delta\in \Delta$ is bounded by the attacker in its $\ell_p$ norm~\cite{shafahi_poison_2018} (\eg, $\|\vct \delta\|_2 \leq \epsilon$). Such box constraint can also be implemented as a soft constraint, as originally done by~\citet{shafahi_poison_2018}.\footnote{The original formulation of feature collision in \cite{shafahi_poison_2018} adopts the $\ell_p$ constraint as soft constraint up-weighted by a Lagrangian penalty term $\beta$, which is basically equivalent to our hard-constraint formulation for  appropriate choices of $\beta$ and $\epsilon$.}
Similarly, \citet{guo_practical_2020} adopted \textit{feature collision} to stage the attack, but they extended the attack's objective function to further increase the poisoning effectiveness. %\textcolor{red}{Differently from the Eq.~\eqref{eq:feature_collision}, \citet{guo_practical_2020} also introduced a soft constraint on the feature representation of the poisoning samples enforcing their separability with respect to the original class of $\vct x$.}
Nevertheless, although this strategy turns out to be effective, it assumes that the feature extractor is fixed and that it is not updated during the fine-tuning process. Moreover, StringRay~\cite{suciu2018does}, ConvexPolytope~\cite{zhu_transferable_2019}, and BullseyePolytope~\cite{aghakhani20bullseye} observed that when reducing the attacker's knowledge the poisoning effectiveness decreases. 
These works showed that \textit{feature collision} is not practical if the attacker does not know exactly the details of the feature extractor, as the embedding of poisoning samples may not be consistent across different feature extractors. To mitigate these difficulties, ConvexPolytope~\cite{zhu_transferable_2019} and BullseyePolytope~\cite{aghakhani20bullseye} optimize the poisoning samples against \textit{ensemble models}, constructing a convex polytope around the target samples to enhance the effectiveness of the attack. The underlying idea is that constructing poisoning samples against ensemble models may improve the attack transferability.
The authors further optimize the poisoning samples by establishing a strong connection among all the layers and the embeddings of the poisoning samples, partially overcoming the assumption that the feature extractor $\phi$ remains fixed. 

All these approaches have the property of creating clean-label samples, as first proposed in \citet{shafahi_poison_2018}, to stay undetected even when the class labels of training points are validated by humans. This is possible as these attacks are staged against deep models, since for these models, small (adversarial) perturbations of samples in the input space correspond to large changes in their feature representations.

\subsubsection{Bilevel Poisoning (Clean-Label)}~\label{sec:targeted_optimized_clean_label}
Although \textit{feature collision} attacks are effective, they may not result in the optimal accuracy, and they do not minimize  the number of poisoned points to change the model's prediction on a single test point.
Moreover, they assume that the training process is not significantly changing the feature embedding. Indeed, when the whole model is trained from scratch, these strategies may not work properly as poisoning samples can be embedded differently. Recent developments, including MetaPoison~\cite{huang_metapoison_2020} and the work by \citet{geiping_witches_2020}, tackle the targeted poisoning attack in the \emph{training-from-scratch} (TS) scenario, while ensuring the clean-label property. These approaches are derived from the bilevel formulation in Eqs.~(\ref{eq:targeted_outer_problem})-(\ref{eq:targeted_inner_problem}), but they exploit distinct and more scalable approaches to target DNNs, and optimize the attack directly against the test samples \fullTest as done in \cite{koh_understanding_2017}.
More concretely, MetaPoison~\cite{huang_metapoison_2020} uses a meta-learning algorithm, as done by \citet{munoz-gonzalez_towards_2017}, to decrease the computational complexity of the attack. They further enhance the transferability of their attack by optimizing the poisoning samples against an ensemble of neural networks, trained with different hyperparameter configurations and algorithms (\eg, weight initialization, number of epochs). 
\citet{geiping_witches_2020} craft poisoning samples to maximize the alignment between the inner loss and the outer loss in Eqs.~\eqref{eq:targeted_outer_problem}-\eqref{eq:targeted_inner_problem}. The authors observed that matching the gradient direction of malicious examples is an effective strategy for attacking DNNs trained from scratch, even on large training datasets. Although modern \textit{feature collision} or optimized strategies are emerging with notable results for targeted attacks, their performance, especially in black-box settings, still demands further investigation.

\subsection{Backdoor (Integrity) Poisoning Attacks}~\label{sec:backdoor_poisoning}
Backdoor poisoning attacks aim to cause an integrity violation. In particular, for any test sample containing a specific pattern, \ie, the so-called \textit{backdoor trigger}, they aim to induce a misclassification, without affecting the classification of clean test samples. The backdoor trigger is clearly known only to the attacker, making it challenging for the defender to evaluate whether a given model provided to them has been backdoored during training or not. 
In Fig.~\ref{fig:backdoor_conceptual} we consider the case where the attacker provides a backdoored street-sign detector that has good accuracy for classifying street signs in most circumstances. However, the classifier has successfully learned the backdoor data distribution, and will output speed-limit predictions for any stop-sign containing the backdoor trigger.
In the following sections, we describe backdoor attacks following the categorization given in Table~\ref{table:poisoning_categorization}. Notably, such attacks have been initially staged in the \emph{model-training} (MT) setting, assuming that the user outsources the training process to an untrusted third-party service, but they have been then extended also to the \emph{training-from-scratch} (TS) and \emph{fine-tuning} (FT) scenarios. 

\subsubsection{Trigger Poisoning} Earlier work in backdoor attacks considered three main families of backdoor triggers, i.e., \textit{patch}, \textit{functional}, and \textit{semantical} triggers, as discussed below.

\textitparagraph{Patch.}~\label{sec:backdoor_patch}
The first threat vector of attack for backdoor poisoning has been investigated in BadNets~\cite{gu_badnets_2017}. The authors considered the case where the user outsources the training process of a DNN to a third-party service, which maliciously alters the training dataset to implant a backdoor in the model.
To this end, the attacker picks a random subset of the training data, blends the backdoor trigger into them, and changes their corresponding labels according to an attacker-chosen class.
A similar idea has been investigated further in LatentBackdoor~\cite{yao_latent_2019} and  TrojanNN~\cite{liu_trojaning_2018}, where the backdoor trigger is designed to maximize the response of selected internal neurons, thus reducing the training data needed to plant the trigger. 
Additionally, LatentBackdoor~\cite{yao_latent_2019} designed the trigger to survive even if the last layers are fine-tuned with novel clean data, while TrojanNN~\cite{liu_trojaning_2018} does not need access to the training data as a reverse-engineering procedure is applied to create a surrogate dataset.
All these attacks assume that the trigger is always placed in the same position, limiting their application against specific defense strategies~\cite{soremekun2020exposing,bajcsy2021baseline,chen2018detecting}. To overcome this issue, BaN~\cite{dynamic_backdoor_salem_2020} introduced different backdoor attacks where the trigger can be attached in various locations of the input image. The underlying idea was to force the model to learn the backdoor trigger and make it location invariant.

\textitparagraph{Functional.}~\label{sec:backdoor_functional}
The patch strategy is based on the idea that poisoning samples repeatedly present a fixed pattern as a trigger, which may however be detected upon human validation of training samples (in the TS and FT scenarios, at least). 
In contrast, a functional trigger represents a stealthier strategy as the corresponding trigger perturbation is slightly spaced throughout the image or changes according to the input. Some works assume to slightly perturb the entire image so that those small variations are not detectable by humans, but evident enough to mislead the model.
In WaNET~\cite{WaNet_nguyen_2021} warping functions are used to generate invisible backdoor triggers (see Fig.~\ref{fig:strategy}). Moreover, the authors enforced the model to distinguish the backdoor warping functions among other pristine ones.
In \citet{steganography_shaofeng_2020} \textit{steganography} algorithms are used to hide the trigger into the training data. Specifically, the attacker replaces the least significant bits to contain the binary string representing the trigger. 
In DFST~\cite{Cheng2021DFST} style transfer generative models are exploited to generate and blend the trigger. 
However, the aforementioned poisoning approaches assume that the attacker can change the labeling process and that no human inspection is done on the training data. 
This assumption is then relaxed by \citet{barni_clean_label_2019} and \citet{ reflection_backdoor_liu_2020}, where clean-label backdoor poisoning attacks are considered; in particular, \citet{reflection_backdoor_liu_2020} used natural reflection effects as trigger to backdoor the system, while \citet{barni_clean_label_2019} used an invisible sinusoidal signal as backdoor trigger (see Fig.~\ref{fig:strategy}). 
More practical scenarios, where the attacker is assumed to have limited knowledge, have been investigated by \citet{chen_targeted_2017} and \citet{invisible_zhong_2020}.
In these two works, the authors used the idea of blending fixed patterns to backdoor the model. In the former approach, \citet{chen_targeted_2017} assume that the attacker blends image patterns into the training data and tunes the blend ratio to create almost invisible triggers, while impacting the backdoor's effectiveness. In the latter, \citet{invisible_zhong_2020} assume that an invisible grid pattern is generated to increase the pixel's intensity, and  its effectiveness is tested in the TS and FT settings.

\textitparagraph{Semantical.}~\label{sec:backdoor_semantical}
The semantical strategy incorporates the idea that backdoor triggers should be feasible and stealthy.
For example, \citet{facehack_sarkar_2020} used facial expressions or image filters (e.g., old-age, smile) as backdoor triggers against real-world facial recognition systems. At training time, the backdoor trigger is injected into the training data to cause the model to associate a smile filter with the authorization of a user. At test time, the attacker can use the same filter to mislead classification. Similarly, \citet{chen_targeted_2017} and \citet{wenger_backdoor_2020} tried to poison face-recognition systems by blending physically-implementable objects (e.g., sunglasses, earrings) as triggers. 

\subsubsection{Bilevel Poisoning}~\label{sec:backdoor_optimized_functional}
Trigger-based strategies assume that the attacker uses a predefined perturbation to mount the attack. However, an alternative strategy for the attacker is to learn the trigger/perturbation itself to enhance the backdoor effectiveness. To this end, even backdoor poisoning can be formalized as a bilevel optimization problem:
\begin{eqnarray}
    \label{eq:backdoor_outer_problem}
    %\min_{\vct \delta} &&  \LossSymb(\cleanVal \cup \targetVal^{\vct t}, \vct\theta^\star) \, ,\\
    \min_{\vct \delta \in \Delta} &&  \LossSymb(\cleanVal, \modelSymb, \vct\theta^\star) + \LossSymb(\targetVal^{\vct t}, \modelSymb, \vct\theta^\star) \, ,\\
    \label{eq:backdoor_inner_problem}
    \text{s.t.} && \vct\theta^\star \in  \argmin_{\vct \theta} \, \RegularizedLossSymb(\cleanTrain \cup \poisonTrain^{\vct \delta},\modelSymb, \vct\theta) \,  .
\end{eqnarray}
%with $(x_k, y_k)\in \cleanVal$, $(x_t, y_t)\in \targetVal$, $(x_c, y_c)\in 

Here, the attacker optimizes the training perturbation $\vct{\delta}$ for poisoning samples in \poisonTrain to mislead the model's prediction for validation samples $\targetVal$ containing the backdoor trigger $\vct t$.
In contrast to indiscriminate and targeted attacks (in Sect.~\ref{sec:availability_poisoning} and Sect.~\ref{sec:targeted_poisoning}), the attacker injects the backdoor trigger in the validation samples $\vct{t}$ to cause misclassifications. Additionally, as for targeted poisoning, the error on \cleanVal is minimized to preserve the system's functionality.

One way to address this bilevel formulation is to craft optimal poisoning samples using generative models~\cite{input_aware_nguyen_2020,Doan_2021_ICCV,Li2021Invisible}, as also done in \cite{yang_generative_2017} for indiscriminate poisoning. \citet{input_aware_nguyen_2020} trained the generative model with a loss that enforces the \textit{diversity} and \textit{noninterchangeable} of the trigger, while LIRA~\cite{Doan_2021_ICCV}'s generator is trained to enforce effectiveness and invisibility of the triggers. 
Conversely, \citet{Li2021Invisible} used a generative neural network steganography technique to embed a backdoor string into poisoning samples.
Another way is to perturb training samples with adversarial noise, as done by \citet{steganography_shaofeng_2020} and \citet{invisible_zhong_2020}. More concretely, in the former approach, the trigger maximizes the response of specific internal neurons, and a regularization term is introduced in the objective function to make the backdoor trigger invisible. In the latter work, the attacker looks for the minimum  universal perturbation that pushes any input towards the decision boundary of a target class. The attacker can use this invisible perturbation trigger on any image, inducing the model to misclassify the target class.

\subsubsection{Feature Collision (Clean-Label)}~\label{sec:backdoor_fc_patch} The backdoor trigger visibility influences the stealthiness of the attack. A backdoor trigger that is too obvious can be easily spotted when the dataset is inspected \cite{saha_hidden_2019}. 
However, Hidden Trigger~\cite{saha_hidden_2019} introduced the idea of using the \textit{feature collision} strategy, seen in Sect.~\ref{sec:targeted_featurecollision} and formulated in Eq.~\eqref{eq:feature_collision}, to hide the trigger into natural target samples.
Specifically, the attacker first injects a random patch trigger into the training set, and then each poisoning sample is masked via \textit{feature collision}. The resulting poisoning images are visually indistinguishable from the target, and have a consistent label (\ie, they are clean-label), while the test samples with the patch trigger will collide with the poisoning samples in feature space, ensuring that the attack works as expected. 

Although the work in \cite{saha_hidden_2019} implements an effective and stealthy clean-label attack, it is applicable only in the feature extractor $\phi$ is not updated. Such a limitation is mitigated by~\citet{turner_clean-label} who exploit a surrogate latent space, rather than $\phi$, to interpolate the backdoor samples, hiding the training-time trigger. Moreover, the attacker can tune the trigger visibility at test time to enhance the attack's effectiveness.
%\citet{turner_clean-label}  were  the first to introduce a clean-label backdoor poisoning attack in the \emph{training-from-scratch} scenario, for which the training and the test trigger are different. 
%Specifically, they used latent space interpolations and adversarial perturbations to create the backdoor samples, and tune the trigger visibility to make the attack less visible during training. During inference, the attacker can use the fully visible trigger to increase its effectiveness. 

\subsubsection{Bilevel Poisoning (Clean-Label)}~\label{sec:backdoor_optimized_patch}
%Subsequently, \citet{sleeper_goldstein_2021} were  the first to introduce a clean-label backdoor poisoning attack in the \emph{training-from-scratch} scenario, for which the training and the test trigger are different. 
%The authors exploit the gradient-alignment technique, used also for targeted attacks in \cite{geiping_witches_2020}, to create effective backdoor poisoning attacks. In both cases, the main advantage is that the backdoor trigger is partially (or even completely~\cite{sleeper_goldstein_2021}) hidden in the training data, and used only at test time.
Inspired by recent success of the gradient-alignment technique in~\cite{geiping_witches_2020} for targeted poisoning, \citet{sleeper_goldstein_2021} exploited the same bilevel-descending strategy to stage clean-label backdoor poisoning attacks in the \emph{training-from-scratch} scenario. Similarly to \citet{saha_hidden_2019} the training and the test data perturbations are different, enhancing the stealthiness of the attack and making it stronger against existing defenses.

%%%%%%%%%%%%%%%%% END OF BACKDOOR %%%%%%%%%%%%%%%%%%

\subsection{Current Limitations}~\label{sec:attack_challenges}
Although data poisoning has been widely studied in recent years, we argue here that two main challenges are still hindering a thorough development of poisoning attacks.
\subsubsection{Unrealistic Threat Models.}~\label{sec:attack_challenge_unrealistic}
The first challenge we formulate here questions some of the threat models considered in previous work. 
%%%%%%%%%%%% pre rebuttal
%The reason is that such threat models are not well representative of what may happen in many real-world application settings. 
%For example, \citet{preventing_fowl_2021} and \citet{confuse_fengCZ_2019} assume that the attacker controls almost the entire training dataset to effectively mount an indiscriminate poisoning attack against DNNs. While this may happen in certain hypothesized situations, it is also not quite surprising that a poisoning attack works if the attacker controls a large fraction of the training set. We believe that poisoning attacks should be considered a realistic threat only when it is assumed that a small fraction of the training points can be controlled by the attacker. We refer the reader to a similar discussion in the context of poisoning federated learning in~\cite{shej22-sp}.
%%%%%%%%%%%% post rebuttal
The reason is that such threat models are not well representative of what may happen in many real-world scenarios for the attackers. They are valuable because they allow system designers to test the system's robustness under worst-case scenarios, but their practicability and effectiveness against realistic production systems are unknown. To give an accurate estimate of how poisoning attacks are effective against ML production systems, we should consider assumptions that are less favorable to the attacker.
For example, \citet{preventing_fowl_2021} and \citet{confuse_fengCZ_2019} assume that the attacker controls almost the entire training dataset to effectively mount an indiscriminate poisoning attack against DNNs. While this may happen in certain hypothesized situations, it is also not quite surprising that a poisoning attack works if the attacker controls a large fraction of the training set. 
%%%%%%%%%%%% pre rebuttal
%We believe that poisoning attacks should be considered a realistic threat only when it is assumed that a small fraction of the training points can be controlled by the attacker. We refer the reader to a similar discussion in the context of poisoning federated learning in~\cite{shej22-sp}.
%%%%%%%%%%%% post rebuttal
We believe that poisoning attacks that assume that only a small fraction of the training points can be controlled by the attacker are more realistic and, therefore, viable against real production systems. We refer the reader to a similar discussion in the context of federated learning poisoning in~\cite{shej22-sp}.

Another limitation of threat models considered for poisoning attacks is that, in some cases, exact knowledge of the \textit{test} samples is implicitly assumed. For example, \cite{shafahi_poison_2018} and \cite{geiping_witches_2020} optimize a targeted poisoning attack to induce misclassification of few specific \textit{test} samples. In particular, the attack is both optimized and tested using the same \textit{test} samples, differently from work which optimizes the poisoning samples using validation data, and then tests the attack impact on a separate test set~\citep{biggio2012poisoning,munoz-gonzalez_towards_2017}.
This evaluation setting clearly enables the attack to reach higher success rates, but at the same time, there is no guarantee that the attack will generalize even to minor variations of the considered test samples, questioning its applicability outside of settings in which the attacker has exact knowledge of the test inputs. For instance, the attack may not work as expected in physical domains, where images are acquired by a camera under varying illumination and environmental conditions. In such cases, it is indeed clear that the attacker can not know beforehand the specific realization of the test sample, as they do not control the acquisition conditions. 
%This assumption, together with the considered setting (\eg, fine-tuning), seems to be helpful in developing heuristics that speed up the implementation of the attack. 
%These works can effectively work in digital scenarios, where the attacker can forward inputs to the model that are the same as the target samples. 
On a similar note, only a few studies on backdoor poisoning have considered real-world scenarios where external factors (such as lighting, camera orientation, etc.) can alter the trigger. Indeed, as done in \cite{shafahi_poison_2018} and \cite{geiping_witches_2020}, most papers consider digital applications where the implanted trigger is nearly unaltered. 

In conclusion, although some recent works seem to have improved the effectiveness of poisoning attacks, their assumptions are often not representative of the actual production system or the attacker's settings, limiting their applicability only in the proposed context.

\subsubsection{Computational Complexity of Poisoning Attacks}~\label{sec:attack_challenge_scalability}
The second challenge we discuss here is related to the solution of the bilevel programming problem used to optimize poisoning attacks. The problem, as analyzed by \citet{munoz-gonzalez_towards_2017}, is that solving the bilevel formulation with a gradient-based  approach requires computing and inverting the Hessian matrix associated to the equilibrium conditions of the inner learning problem, which scales cubically in time and quadratically in space with respect to the number of model's parameters. Even if one may exploit rank-one updates to the Hessian matrix, and Hessian-vector products coupled with conjugate descent to speed up the computation of required gradients, the approach remains too computationally demanding to attack modern deep models, where the number of parameters is on the order of millions.
%Although as anticipated by \citet{munoz-gonzalez_towards_2017} the bilevel approach turns out to be computationally expensive, it is also true that that strategy improves the effectiveness of the attack and its stealthiness against defenses.
Nevertheless, it is also true that that solving the bilevel problem is expected to improve the effectiveness of the attack and its stealthiness against defenses.
%Targeted
For example, the bilevel strategy approach is the only one at the state of the art which allows mounting an effective attack in the \emph{training-from-scratch} (TS) setting. Other heuristic approaches, e.g., \textit{feature collision}, have been shown to be totally ineffective if the feature extractor $\phi$ is updated during training \cite{geiping_witches_2020}.
For backdoor poisoning, the recent developments in the literature show that bilevel-inspired attacks are more effective and can better counter existing defenses \cite{Doan_2021_ICCV,sleeper_goldstein_2021,input_aware_nguyen_2020}.
Thus tackling the complexity of the bilevel poisoning problem remains a relevant open challenge to ensure a fairer and scalable evaluation of modern deep models against such attacks.

\subsection{Transferability of Poisoning Attacks}~\label{sec:transferability}
Transferability is a characteristic of attacks to be effective even against classifiers the attacker does not have full knowledge about. 
The term transferability was first investigated for adversarial examples in~\cite{goodfellow15explaining,papernot17-asiaccs,Papernot16transferability}. In case of limited knowledge (\ie, black-box attacks), the attacker can use surrogate learners or training data to craft the attack and transfer it to mislead the unknown target model. 
Nevertheless, the first to introduce the idea of surrogates for data poisoning attacks were \citet{nelson08spam} and \citet{biggio2012poisoning}. The authors claimed that if the attacker does not have exact knowledge about the training data, they could sample a surrogate dataset from the same distribution and transfer the attack to the target learner.
In subsequent work, \citet{munoz-gonzalez_towards_2017} and \citet{demontis2019whytransferability} analyzed the transferability of poisoning attacks using also surrogate learners, showing that matching the complexity of the surrogate and the target model enhances the attack effectiveness.
Transferability has also been investigated when considering surrogate objective functions. More concretely, optimizing attacks against a smoother objective function may find effective, or even better, local optima than the ones of the target function~\cite{private_learners_ma_2019,demontis2019whytransferability,koh_understanding_2017,Papernot16transferability}. For example, optimizing a non-differentiable loss can be harder, thus using a smoothed version may turn out to be more effective~\cite{koh_understanding_2017}.
More recently, \citet{suciu2018does} showed that the attacker can leverage transferability even when the attacker has limited knowledge about the feature representation, at the cost of reducing the attack effectiveness. 
However, \citet{zhu_transferable_2019} and \citet{aghakhani20bullseye} independently hypothesize that the stability of \textit{feature collision} attacks is compromised when the feature representation in the representation space is changed. To mitigate this problem, they craft poisoning samples to attack an ensemble of models, encouraging their transferability against multiple networks.

\subsection{Unifying Framework}~\label{sec:unifying_framework}
Although the three poisoning attacks are detailed in Sects.~\ref{sec:availability_poisoning}-\ref{sec:backdoor_poisoning} aim to cause different violations, they can be described by the following generalized bilevel programming problem:
\begin{comment}
\begin{eqnarray}
    \label{eq:poisoning_outer_problem}
    \max_{\vct\delta} && \alpha \textstyle \sum_{k=1}^{N_k} \lossSymb(\vct{x}_k, y_k, \vct\theta^\star) + \beta \sum_{t=1}^{N_t}\lossSymb(\vct{x}_t + \vct{t}_t, \poisonLabel_t, \vct\theta^\star) \, ,\\
    \label{eq:poisoning_inner_problem}
   {\rm s.t.}  && \vct\theta^\star \in  \argmin_{\vct\theta} \quad \textstyle \sum_{c=1}^{N_c} \lossSymb(\vct{x}_c, y_c, \vct\theta) + \sum_{p=1}^{N_p} \lossSymb(\vct x_p+ \vct{\delta}_p, \poisonLabel_p, \vct\theta) \, ,
\end{eqnarray}
\end{comment}
\begin{eqnarray}
    \label{eq:poisoning_outer_problem}
    \max_{\vct \delta \in \Delta} &&  \alpha\LossSymb(\cleanVal,  \modelSymb, \vct\theta^\star) - \beta \LossSymb(\targetVal^{\vct t},  \modelSymb, \vct\theta^\star) \, ,\\
    \label{eq:poisoning_inner_problem}
    \text{s.t.} && \vct\theta^\star \in  \argmin_{\vct \theta} \, \RegularizedLossSymb(\cleanTrain \cup \poisonTrain^{\vct \delta},  \modelSymb, \vct\theta) \,  ,
\end{eqnarray}
%The optimization program in Eqs.~\eqref{eq:poisoning_outer_problem}-\eqref{eq:poisoning_inner_problem} aims to accomplish the attacker's goal, considering its capacity and knowledge, by optimizing the perturbation $\vct \delta$ used to poison the training samples \poisonTrain. 
The optimization program in Eqs.~\eqref{eq:poisoning_outer_problem}-\eqref{eq:poisoning_inner_problem} aims to accomplish the attacker's goal, considering their capacity of tampering with the training set and knowledge of the victim model, by optimizing the perturbation $\vct \delta$ used to poison the training samples in \poisonTrain. Additionally, as in Eqs.~\eqref{eq:dos_outer_problem}-\eqref{eq:backdoor_inner_problem}, the poisoning noise $\vct \delta$ belongs to $\Delta$ which encompass possible domain constraints or feature constraints to improve stealthiness of the attack (\eg, invisibility of the trigger).
% Before rebuttal
%The test data perturbation $\vct t$ is either absent \textcolor{red}{(\ie, $\vct t = \vct 0$)}, for indiscriminate and target poisoning, or pre-defined/optimized by the adversary for backdoor poisoning. In other words, differently from adversarial examples~\cite{goodfellow15explaining,biggio13evasion}, the perturbation $\vct t$ is not optimized at test time.
%After rebuttal 
The test data perturbation $\vct t$ is absent (\ie, $\vct t = \vct 0$), for indiscriminate and target poisoning. For backdoor poisoning, $\vct t$ is pre-defined/optimized by the attacker before training, unlike from adversarial examples~\cite{goodfellow15explaining,biggio13evasion} where the perturbation $\vct t$ is optimized at test time.
The coefficients $\alpha$ and $\beta$ are calibrated according to the attacker's desired violation. We can set: (i) $\alpha=1 (-1)$ and $\beta=0$ for error-generic (specific) indiscriminate poisoning; (ii) $\alpha=-1$ and $\beta=-1 (1)$ for error-specific (generic) targeted poisoning; (iii) $\alpha=-1$ and $\beta=-1 (1)$ for error-specific (generic) backdoor poisoning. 

In conclusion, although backdoor, indiscriminate and targeted attacks are designed to cause distinct security violations, they can be formulated under a unique bilevel optimization program. Therefore, as we will explore in Sec.~\ref{sec:development}, solutions for optimizing bilevel optimization programs fast can pave the way towards developing novel effective and stealthy poisoning attacks capable of mitigating the scalability limit of current strategies.

\section{Defenses}~\label{sec::defenses}
Many defenses have been proposed to mitigate poisoning attacks.
In this section, we discuss each of the six defense classes identified in Sect.~\ref{sec::defenseSetting}. For each group, we review the related learning and defense settings, and the various approaches suggested by prior works. Some defenses can be assigned to several groups. In these cases, we assigned a defense to the most suitable group in terms of writing flow. A compact summary of all defenses is given in Table~\ref{table:defensesAll}. We further match attack strategies and defenses at training and test time in Table~\ref{table:matching_attack_defenses}. Having reviewed all defense groups, we conclude the section by discussing current defense evaluation issues, outlining three main open challenges.

\begin{table*}[t]
    \centering
    \caption{Overview of defenses in the area of classification. 
    When several approaches are named, we order them according to publication year and alphabetical order of the authors.
    %For capabilities and knowledge, \yes \ indicates that the knowledge or capability is present, \no \ that it is not. In $\theta$, \grayc \ refers to the ability to fine-tune the model. We further denote whether the approach provides a certification (Cert.) or is based on a DNN.
    For each paper we report the defender's knowledge and capability required, consisting in access to the training data \fullTrain, clean validation data \cleanVal and access to the training procedure \learningSymb and the model's parameters $\theta$. \yes \ indicates that the corresponding knowledge or capability is present, \no \ that it is not. In $\theta$, \grayc \ refers to the ability to fine-tune the model. We further denote whether the approach provides a certification (Cert.), or it is suited to deep neural networks (DNN). $^*$~intended as forensic tool to determine which points were poisoned in retrospect, not as a defense.}
    \resizebox{\textwidth}{!}{%
      \renewcommand{\arraystretch}{0.93}
    \begin{tabular}{lllcccccc}
        \toprule
%        & &  \multicolumn{3}{r}{Defender's knowledge}& \multicolumn{4}{l}{and capabilities} \\
%        \cmidrule(r){4-5}  \cmidrule(r){6-7}  
 & Defense strategy & Defense & $\fullTrain$ & $\cleanVal$ & \learningSymb & $\vct \theta $ & Cert. & DNN \\ 
 \midrule 
 %%%% availability
 \multicolumn{1}{l}{\multirow{9}{*}{\rotvertical{Indiscriminate}}}  & \multicolumn{1}{l}{\multirow{3}{*}{\ref{sec:trainingDataSanitization} Training Data Sanitization}}\rdelim\{{3}{*}  & Taheri et al.~\cite{taheri2020defending} & \yes & \yes & \no & \no & & \\ %Malware
%Casting out Demons~\cite{Cretu2008Casting-out-Dem} & \yes & \no & \yes & \yes & &  \\ %
%& & Taheri et al.~\cite{taheri2020defending} & \yes & \yes & \no & \no & & \\ %regression
&&  Curie~\cite{DBLP:journals/corr/LaishramP16}, Paudice et al.~\cite{paudice2018label}, \citet{frederickson2018attack} & \yes & \no & \no & \no & &\\
& & Sphere / Slab Defense~\cite{2017arXiv170603691S} & \yes  & \no  & \no  & \no   & \checkmark & \\ 
\addlinespace
\multicolumn{1}{l}{}  & \multicolumn{1}{l}{\multirow{6}{*}{\ref{sec:robustTraining} Robust Training}}\rdelim\{{6}{*}    &  RONI~\cite{nelson2009misleading}, \citet{biggio11labelnoise},  \citet{DBLP:conf/itasec/DemontisBFGR17}
   & \multirow{1}{*}{\yes}  & \multirow{1}{*}{\no}  & \multirow{1}{*}{\yes}  &\multirow{1}{*}{\yes}   &  \\ 
     &   & Sever~\cite{diakonikolas2019sever}, Jia et al.~\cite{jia2020certified}, \citet{rosenfeld2020certified} & \yes  & \no  & \yes  & \yes   &\checkmark  & \\ 
  & & \citet{hong2020effectiveness} & \yes & \no & \yes & \yes & & \checkmark \\
   &   & (SS-)DPA~\cite{levine2020deep} & \yes  & \no  & \no  & \yes   &\checkmark  & \checkmark \\ 
 \multicolumn{1}{l}{}  &   &  Weighted Bagging~\cite{biggio2011bagging} & \yes & \no & \no & \yes  & \\
  \multicolumn{1}{l}{}  &   &  Diff. Private Learners~\cite{private_learners_ma_2019}, \citet{Chen2022Collective} \citet{Wang2022Improved} & \yes & \no & \no & \yes & \checkmark \\
 
 \addlinespace
%\addlinespace
%\multicolumn{1}{l}{\multirow{1}{*}{\rotvertical{Both}}}  & robust training  & Gradient Shaping~\cite{hong2020effectiveness} & \yes  & \no  & \yes  & \yes   &  \\ 
%\addlinespace
%\addlinespace
 
%%%%% INTEGRITY
  %%%
  \addlinespace
%  \multicolumn{1}{l}{\multirow{24}{*}{\rotvertical{Backdoor / Targeted}}} & \multicolumn{1}{l}{\multirow{2}{*}{\ref{sec:trainingDataSanitization} Training Data Sanitization}} \rdelim\{{2}{*} & Spectral Signatures~\cite{tran2018spectral},  CI~\cite{xiang2019benchmark}, \citet{peri2020deep} &\multirow{2}{*}{\yes}  & \multirow{2}{*}{\no}  & \multirow{2}{*}{\no}  &\multirow{2}{*}{\no}   & & \multirow{2}{*}{\checkmark} \\ 
  \multicolumn{1}{l}{\multirow{17}{*}{\rotvertical{Backdoor / Targeted}}} & \multicolumn{1}{l}{\multirow{3}{*}{\ref{sec:trainingDataSanitization} Training Data Sanitization}} \rdelim\{{3}{*} & \citet{shawn2022traceback}$^*$ & \yes & \no & \yes & \yes && \checkmark \\
  && Spectral Signatures~\cite{tran2018spectral},  CI~\cite{xiang2019benchmark}, \citet{peri2020deep} &\multirow{2}{*}{\yes}  & \multirow{2}{*}{\no}  & \multirow{2}{*}{\no}  &\multirow{2}{*}{\no}   & & \multirow{2}{*}{\checkmark} \\ 
  & & RE~\cite{xiang2021reverse}, SPECTRE~\cite{hayase2021spectre} &  &   &   &  & \\ 
  \addlinespace
& \multicolumn{1}{l}{\multirow{3}{*}{\ref{sec:robustTraining} Robust Training}}\rdelim\{{3}{*}   & \citet{du2019diffprivacy}, \citet{hong2020effectiveness},  \citet{borgnia2021strong},  \citet{geiping2021doesn} & \multirow{2}{*}{\yes}  & \multirow{2}{*}{\no}  & \multirow{2}{*}{\yes}  & \multirow{2}{*}{\yes}   &  &\multirow{2}{*}{\checkmark} \\ 
& & ABL~\cite{li2021anti}, \citet{huang2021decouopling}, \citet{sun2021can}, \citet{yo2022CreatedEqual} \\

  %&  & Borgnia et al.~\cite{borgnia2021strong}, Geiping et al.~\cite{geiping2021doesn} & \yes  & \no  & \yes  & \yes  & \checkmark  &  & \\ 
  &  &  DP-InstaHide~\cite{borgnia2021dp},  RAB~\cite{weber2020rab} & \yes  & \no  & \yes  & \yes   &\checkmark  & \checkmark \\ 
  %ABL~\cite{li2021anti} & \yes & \no & \yes & \yes & & \checkmark
  
  \addlinespace
  & \multicolumn{1}{l}{\multirow{5}{*}{\ref{sec:modelInspection} Model Inspection}}\rdelim\{{5}{*}   &   AEGIS~\cite{soremekun2020exposing} & \no & \yes & \yes & \yes  & & \checkmark\\
&   & Neuroninspect~\cite{huang2019neuroninspect}, \citet{bajcsy2021baseline}, L-RED~\cite{xiang2021red} & \no  & \yes  & \no  & \yes   &  & \checkmark\\ 
  %&   & Neuroninspect~\cite{huang2019neuroninspect} & \no  & \yes  & \no  & \yes    &   \\
    & & Activation Clustering~\cite{chen2018detecting}, \citet{tang2019demon} & \yes  & \no  & \no  & \no   &  & \checkmark\\ 
  %&   & Activation Clustering~\cite{chen2018detecting} & \yes  & \no  & \no  & \no    &  \\ 
  & & MNTD~\cite{Xu2021MetaAudio}, Litmus patterns~\cite{kolouri2020universal}, AEVA~\cite{guo2022aeva} & \no  & \yes  & \no  & \no  & & \checkmark   \\ 
&   & DeepInspect~\cite{chen2019deepinspect} & \no  & \no  & \no  & \no  & & \checkmark  \\  
  \addlinespace
& \multicolumn{1}{l}{\multirow{6}{*}{\ref{sec:modelSanitization} Model Sanitization}}\rdelim\{{6}{*}   &  I-BAU~\cite{zeng2021adversarial} & \no & \yes & \yes & \yes && \checkmark \\
&& Yoshida et al.~\cite{yoshida2020disabling} & \yes  & \no  & \yes  & \grayc    &  & \checkmark \\
  &  & Cheng et al.~\cite{cheng2020defending}, Zhao et al.~\cite{zhao2020bridging},  ANP~\cite{wu2021adversarial},  CLEAR~\cite{zhu2021clear} & \no  & \yes  & \no  & \yes    &  & \checkmark \\ 
    &  & Re-training~\cite{DBLP:conf/iccd/LiuXS17} & \yes  & \no  & \no  & \yes   &  & \checkmark\\ 
& & Liu et al.~\cite{liu2018fine}, Neural Attention Distillation~\cite{li2021neural} & \no  & \yes  & \no  & \grayc  & & \checkmark \\ 

    %&  & Neural Attention Distillation~\cite{li2021neural} & \no  & \yes  & \no  & \grayc   &   \\ 
    &   & DeepSweep~\cite{zeng2020deepsweep} & \yes  & \no  & \no  & \grayc  & & \checkmark \\  
   \addlinespace    
   \addlinespace
    \multicolumn{1}{l}{\multirow{11}{*}{\rotvertical{Backdoor}}}  & \multicolumn{1}{l}{\multirow{5}{*}{\ref{sec:triggerReconstruction} Trigger Reconstruction}}\rdelim\{{5}{*}  & ABS~\cite{liu2019abs}, Neural Cleanse~\cite{wangneural}, Shen et al.~\cite{shen2021backdoor} & \multirow{2}{*}{\no}  & \multirow{2}{*}{\yes}  & \multirow{2}{*}{\no}  & \multirow{2}{*}{\yes}   & & \checkmark \\   
  %    &   & ABS~\cite{liu2019abs}, Neural Cleanse~\cite{wangneural} & \yes  & \no  & \no  & \yes    &   \\ 
  && NEO~\cite{udeshi2019model}, \citet{hu2022trigger}  \\%& \no  & \yes  & \no  & \yes    & & \checkmark \\ 
  &   & MESA~\cite{qiao2019defending}, Gangsweep~\cite{zhu2020gangsweep}, Xiang et al.~\cite{xiang2020detection} & \multirow{2}{*}{\no}  & \multirow{2}{*}{\yes}  &\multirow{2}{*}{\no}  & \multirow{2}{*}{\no}    & & \multirow{2}{*}{\checkmark} \\ 
  & & TAD~\cite{zhang2021tad}, AEVA~\cite{guo2022aeva},  \citet{xiang2022posttraining} \\
    &   & Tabor~\cite{guo2019tabor}, B3D-SS~\cite{dong2021black} & \no  & \no  & \no  & \no    & & \checkmark   \\ 
 \addlinespace
   & \multicolumn{1}{l}{\multirow{6}{*}{\ref{sec:testInputSanitization} Test Data Sanitization}}\rdelim\{{6}{*}   & NNoculation~\cite{veldanda2020nnoculation} & \no  & \yes  & \yes  & \grayc  &  & \checkmark \\ 
           &  & Anomaly Detection~\cite{DBLP:conf/iccd/LiuXS17} & \yes  & \no  & \no  & \yes  & & \checkmark \\ 
  & & Input Preprocessing~\cite{DBLP:conf/iccd/LiuXS17}, SentiNet~\cite{chou2018sentinet}, ConFoc~\cite{villarreal2020confoc}, & \multirow{2}{*}{\no}  & \multirow{2}{*}{\yes}  & \multirow{2}{*}{\no}  & \multirow{2}{*}{\yes}   & &\multirow{2}{*}{\checkmark} \\ 
  &  & CleaNN~\cite{javaheripi2020cleann}, Februus~\cite{doan2020februus}  &   &   &   &    &   \\ 
     &   & STRIP~\cite{gao2019strip} & \no  & \yes  & \no  & \no   & & \checkmark \\ 
  &  & Li et al.~\cite{li2020rethinking}, Sarkar et al.~\cite{sarkar2020backdoor} & \no  & \no  & \no  & \no  & & \checkmark \\ 
  %%%%%
\bottomrule
    \end{tabular}
    }
    \label{table:defensesAll}
\end{table*}

\begin{table}[t]
\centering
\caption{Matching poisoning attack strategies and defenses. For each defense, we depict on which attack strategy (as defined in Section~\ref{sec::attacks}) the defense was evaluated. We mark cells with \dashmark\xspace if the corresponding defense category have not been investigated so far for the corresponding attack. Conversely, we mark cells with \xmark\xspace if corresponding defense has no sense and cannot be applied.}
\label{table:matching_attack_defenses}
\setlength\tabcolsep{3.1pt}
\renewcommand{\arraystretch}{1.05}
\begin{tabular}{@{}P{0.1cm}P{0.25cm}P{0.25cm}P{0.7cm}|P{1.8cm}P{2.4cm}P{1.8cm}P{1.8cm}P{1.9cm}|P{1.7cm}@{}}
\toprule
\multicolumn{4}{c}{\multirow{2}{*}{Attack}} & \multicolumn{6}{c}{Defenses} \\
\multicolumn{4}{c}{} & \multicolumn{5}{c}{Training Time} & Test Time \\ \midrule
 & $\vct \delta$ & $\vct t$ & \begin{tabular}[c]{@{}c@{}}Clean\\ Label\end{tabular} & \begin{tabular}[c]{@{}c@{}}Training Data\\ Sanitization\end{tabular} & \begin{tabular}[c]{@{}c@{}}Robust \\ Training\end{tabular} & \begin{tabular}[c]{@{}c@{}}Model \\ Inspection\end{tabular} & \begin{tabular}[c]{@{}c@{}}Model \\ Sanitization\end{tabular} & \begin{tabular}[c]{@{}c@{}}Trigger\\ Reconstruction\end{tabular} & \begin{tabular}[c]{@{}c@{}}Test Data\\ Sanitization\end{tabular} \\ \midrule
 %%%indiscriminate
\multirow{3}{*}{\rotvertical{Indiscr.}} & LF & - &  & \cite{paudice2018label, DBLP:journals/corr/LaishramP16,taheri2020defending} & \cite{hong2020effectiveness,levine2020deep,biggio11labelnoise,DBLP:conf/itasec/DemontisBFGR17,rosenfeld2020certified,Wang2022Improved,Chen2022Collective} & \dashmark & \dashmark & \xmark & \xmark \\
 & BL & - &  & \cite{frederickson2018attack,2017arXiv170603691S} & \cite{nelson2009misleading,biggio2011bagging, jia2020certified, private_learners_ma_2019} & \dashmark & \dashmark & \xmark & \xmark \\
 & BL & - & \checkmark & \dashmark & \dashmark & \dashmark & \dashmark & \xmark & \xmark \\ \midrule
 %%%%%%targeted
\multirow{3}{*}{\rotvertical{Targeted}} & BL & - &  & \cite{frederickson2018attack} & \dashmark & \dashmark & \dashmark & \xmark & \dashmark \\
 & FC & - & \checkmark & \cite{peri2020deep,shawn2022traceback,yo2022CreatedEqual} & \cite{geiping2021doesn,borgnia2021dp,li2021anti,hong2020effectiveness} & \cite{zhu2021clear,tang2019demon} & \cite{zhu2021clear} & \xmark & \dashmark \\
 & BL & - & \checkmark & \cite{shawn2022traceback,yo2022CreatedEqual} & \cite{geiping2021doesn,borgnia2021dp,borgnia2021strong} & \dashmark & \dashmark & \xmark & \dashmark \\ \midrule
 %%%%%backdoors
\multicolumn{1}{l}{\multirow{7}{*}{\rotvertical{Backdoor}}} & T$^P$ & T$^P$ &  & \cite{tran2018spectral,xiang2021reverse,xiang2019benchmark,villarreal2020confoc,hayase2021spectre,shawn2022traceback} & \cite{yoshida2020disabling,weber2020rab,borgnia2021strong,li2021anti,jia2020certified,geiping2021doesn,huang2021decouopling,sun2021can,du2019diffprivacy} & \cite{tang2019demon,chen2018detecting,kolouri2020universal,zhang2021tad,shen2021backdoor,chen2019deepinspect,Xu2021MetaAudio,bajcsy2021baseline,xiang2021red,soremekun2020exposing,guo2022aeva,xiang2022posttraining,hu2022trigger} & \cite{liu2018fine,zeng2020deepsweep,yoshida2020disabling,zhao2020bridging,qiao2019defending,chen2019deepinspect,li2021neural,zhu2020gangsweep,wu2021adversarial,zeng2021adversarial} & \cite{guo2019tabor,wangneural,liu2019abs,udeshi2019model,qiao2019defending,cheng2020defending,dong2021black,xiang2021reverse,guo2022aeva,xiang2022posttraining,hu2022trigger} & \cite{doan2020februus,li2020rethinking,zeng2020deepsweep,veldanda2020nnoculation,wangneural,chou2018sentinet,gao2019strip,sarkar2020backdoor,qi2020onion,villarreal2020confoc} \\
\multicolumn{1}{l}{} & T$^S$ & T$^S$ &  & \dashmark & \cite{huang2021decouopling,shawn2022traceback} & \cite{shen2021backdoor,guo2022aeva} &  \cite{DBLP:conf/iccd/LiuXS17,liu2018fine,zeng2020deepsweep,zeng2021adversarial} & \cite{guo2022aeva} &  \cite{DBLP:conf/iccd/LiuXS17,veldanda2020nnoculation}\\
\multicolumn{1}{l}{} & T$^F$ & T$^F$ &  &  \cite{xiang2021reverse,hayase2021spectre} & \cite{weber2020rab,li2021anti,huang2021decouopling} &  \cite{tang2019demon,shen2021backdoor,huang2019neuroninspect,Xu2021MetaAudio,soremekun2020exposing,xiang2021red,hu2022trigger} & \cite{zhu2020gangsweep, zeng2020deepsweep,li2021neural,wu2021adversarial,zeng2021adversarial} &  \cite{xiang2020detection,zhu2020gangsweep,liu2019abs,xiang2021red,xiang2021reverse,hu2022trigger} &  \cite{zeng2020deepsweep,gao2019strip} \\ 
\multicolumn{1}{l}{} & FC & T$^P$ & \checkmark &  \cite{hayase2021spectre} &  \cite{geiping2021doesn,borgnia2021dp,li2021anti,huang2021decouopling,yo2022CreatedEqual} &  \cite{zhu2021clear,guo2022aeva} &  \cite{zhu2021clear,zhu2020gangsweep,li2021neural,wu2021adversarial} & \cite{guo2022aeva} &  \cite{li2020rethinking} \\
\multicolumn{1}{l}{} & BL & T$^F$ &  & \dashmark & \dashmark &  \cite{shen2021backdoor,guo2022aeva} &  \cite{wu2021adversarial} &  \cite{zhu2020gangsweep} &  \cite{zeng2020deepsweep} \\
& BL & T$^P$ & \checkmark & \dashmark&\cite{yo2022CreatedEqual} & \dashmark & \dashmark & \dashmark & \dashmark \\
\bottomrule
\end{tabular}
\end{table}

\subsection{Training Data Sanitization}\label{sec:trainingDataSanitization}
These defenses aim to identify and remove poisoning samples \textit{before training}, to alleviate the effect of the attack. The underlying rationale is that, to be effective, poisoning samples have to be \textit{different} from the rest of the training points. Otherwise, they would have no impact at all on the training process. Accordingly, poisoning samples typically exhibit an outlying behavior with respect to the training data distribution, which enables their detection. The defenses that fall into this category require access to the training data \fullTrain, and in a few cases also access to clean validation data $\cleanVal$, i.e., to an untainted dataset that can be used to facilitate detection of \textit{outlying} poisoning samples in the training set. No capabilities are required to alter the learning algorithm \learningSymb or to train the model parameters $\vct\theta$. Theoretically, these defenses can be applied in all learning settings. We can however not exclude the possibility in the \emph{model-training} setting that the attacker tampers with the data provided, which is beyond the defender's control.
We first discuss defenses against indiscriminate poisoning. 
\citet{paudice2018label} target label-flip attacks by using label propagation.
As \citet{2017arXiv170603691S} show, the difference between poisons and benign data allows to use outlier detection as a defense. Detection can also be eased by taking into account both features and labels, using clustering techniques for indiscriminate~\cite{taheri2020defending,DBLP:journals/corr/LaishramP16} and backdoor/targeted attacks~\cite{shawn2022traceback}. 
Backdoor and targeted poisoning attacks can also be detected using outlier detection, where the outlier is determined in the networks' latent features on the potentially tampered data~\cite{tran2018spectral,hayase2021spectre,peri2020deep}. An orthogonal line of work, by \citet{xiang2019benchmark,xiang2021reverse}, reconstructs the backdoor trigger and removes samples containing it. 
As shown in Table~\ref{table:matching_attack_defenses}, training data sanitization has been applied against various attack strategies. Attack strategies that have not been mitigated yet are only indiscriminate clean-label bilevel attacks, semantical trigger backdoors and bilevel backdoors.

\subsection{Robust Training}\label{sec:robustTraining}
Another possibility to mitigate poisoning attacks is \emph{during training}. 
The underlying idea is to design a training algorithm that limits the influence of malicious samples and thereby alleviates the influence of the poisoning attack.
Intuitively, as reported in Table~\ref{table:defensesAll}, all of these defenses require access to the training data \fullTrain and none to clean validation data $\cleanVal$. Nonetheless, they require altering the learning algorithm \learningSymb and  access to the model's parameters $\vct\theta$. Hence, robust training can only be implemented when the defender trains the model, e.g., in the \emph{training-from-scratch} or \emph{fine-tuning} setting.
To alleviate the effect of indiscriminate poisoning attacks, the training data can be split into small subsets.
The high-level idea is that a larger number of poisoning samples is needed to alter all small classifiers. The defender can build such ensembles using bagging~\cite{levine2020deep,Wang2022Improved,biggio2011bagging} or voting mechanisms~\cite{jia2020certified} or a combination thereof~\cite{levine2020deep,Chen2022Collective}.  An alternative approach by \citet{nelson2009misleading} is to exclude a sample from training if it leads to a significant decrease in accuracy when used in training. 
In addition, \citet{diakonikolas2019sever} apply techniques from robust optimization and robust statistics, thereby limiting the impact of individual, poisonous points. 
Alternatively, the influence of poisons can be limited by increasing the level of regularization~\cite{biggio11labelnoise, DBLP:conf/itasec/DemontisBFGR17}. 
The alleviating effect of regularization against backdoors has been described by~\citet{carnerero2021regularization}, with a more detailed analysis by~\citet{cina2021backdoor}. The latter work  shows that hyperparameters related to regularization affect backdoor performance.
Backdoor and targeted poisoning attacks can also be mitigated using data augmentations like mix-up~\cite{borgnia2021dp,borgnia2021strong}, or based on the model's gradients wrt. the input~\cite{geiping2021doesn}.
%Backdoor and targeted poisoning attacks can also be mitigated using data augmentations like MixUp~\cite{borgnia2021dp,borgnia2021strong} or the adversarial training variant proposed by~\citet{geiping2021doesn}.
Analogously, the data can be augmented using noise to mitigate indiscriminate~\cite{rosenfeld2020certified} and backdoor~\cite{weber2020rab} attacks.
Furthermore, differences in the loss between backdoored/targeted and clean data allows to unlearn~\cite{li2021anti} or identify~\cite{yo2022CreatedEqual} poisons later in training.
 Alternatively, a trained preprocessor can alleviate the threat of backdoors~\cite{sun2021can}. Furthermore, \citet{huang2021decouopling} show that pre-training the network unsupervisedly (e.g., without wrong labels) can alleviate backdoors.
Finally, in both indiscriminate~\cite{hong2020effectiveness,private_learners_ma_2019} and backdoor/targeted~\cite{hong2020effectiveness,borgnia2021dp,du2019diffprivacy} attacks, the framework of differential privacy can be used to alleviate the effect of poisoning. The intuition behind this approach is that differential privacy limits the impact individual data points have, thereby limiting the overall impact of outlying poisoning samples too~\cite{hong2020effectiveness}.
However, further investigation is still required to defend against some bilevel strategies, as visible in Table~\ref{table:matching_attack_defenses}.

\subsection{Model Inspection}\label{sec:modelInspection}
Starting with model inspection,
we discuss groups of defenses operating before the model is deployed. The approaches in these groups mitigate only backdoor and targeted attacks. In model inspection, we determine for a given model whether a backdoor is implanted or not. 
The defense settings in this group are diverse, and encompass all combinations.
In principle, model inspection can be used in all learning settings, where exceptions for specific defenses might apply.
To inspect a model can be formulated as classifications tasks. For example, \citet{kolouri2020universal} and \citet{Xu2021MetaAudio} show that crafting specific input patterns and training a meta-classifier on the outputs of a given model computed on such inputs can reveal whether the model is backdoored. 
\citet{bajcsy2021baseline} 
follow a similar approach, using clean data and a pruned model.
A different observation is that when relying on the backdoor trigger to output a class, the network behaves somehow \emph{unusual}: it will rely on normally irrelevant features.  Thus, outlier detection can be used.
For example,  \citet{zhu2021clear} alternatively search for a set of points that are reliably misclassified to detect feature-collision attacks. 
To detect backdoors and backdoored models, outlier detection can be used on top of interpretability techniques~\cite{huang2019neuroninspect}, or latent representations~\cite{chen2018detecting, tang2019demon,soremekun2020exposing}. 
Alternatively, \citet{xiang2021red} show that finding a trigger that is  reliably misclassified indicates the model is backdoored. 
As reported in Table~\ref{table:matching_attack_defenses}, model inspection has primarily been evaluated on backdoor attacks with a predefined trigger strategy.

\subsection{Model Sanitization}\label{sec:modelSanitization}
Once a backdoored model is detected, the question becomes how to sanitize it. Sanitization requires diverse defense settings encompassing all possibilities.
Model sanitization often involves (re-)training or fine-tuning. Depending on the exact \emph{model-training} setting, sanitizing the model might be impossible (e.g., if the model is provided as a service accessible only via queries). 
To sanitize a model, 
pruning~\cite{wu2021adversarial,cheng2020defending}, retraining~\cite{zeng2021adversarial}, or fine-tuning~\cite{liu2018fine,DBLP:conf/iccd/LiuXS17} can be used. 
Given knowledge of the trigger,
\citet{zhu2021clear} propose to relabel the identified poisoned samples after the trigger is removed.
Alternatively, approaches such as data augmentation~\cite{zeng2020deepsweep} or distillation~\cite{yoshida2020disabling,li2021neural} can augment small, clean datasets. Finally, \citet{zhao2020bridging} show that path connection between two backdoored models, using a small amount of clean data, also reduces the success of the attack.
As shown in Table~\ref{table:matching_attack_defenses}, model sanitization has been evaluated mainly against backdoor attacks. Extensions to other kinds of triggers and targeted attacks might however be possible.

\subsection{Trigger Reconstruction}\label{sec:triggerReconstruction}
As an alternative to model sanitization, this category of defenses aim to reconstruct the implanted trigger.
The assumptions on the defender's knowledge and capabilities are diverse, and encompass many possibilities, although the learning algorithm \learningSymb is never altered. As for model inspection, trigger reconstruction can in theory be used in all learning settings, where exceptions for specific defenses might apply.
While a trigger can be randomly generated~\cite{udeshi2019model,zhang2021tad}, the question remains on how to verify that the reconstructed pattern is a trigger. Many techniques leverage the fact that
a trigger changes the classifier's output reliably.
 This finding has been in detail investigated by 
 \citet{grosse2021backdoor}, who show that backdoor patterns lead to a very stable or smooth output of the target class. In other words, the classifier ignores other features and only relies on the backdoor trigger. 
 Such a stable output also enables to reformulate trigger reconstruction as an optimization problem~\cite{wangneural}. In the first approach of its kind, Wang et al.'s Neural Cleanse~\cite{wangneural}
 optimizes a pattern that leads to reliable misclassification of a batch of input points.
  %uses a batch of input points and searches for a pattern that leads to reliable misclassification of the whole batch.
  The idea is that if there is such a pattern, and it is small, it must be similar to the backdoor trigger. Wang et al.'s  approach has been improved in terms of how to determine whether a pattern is indeed a trigger~\cite{guo2019tabor,xiang2020detection}, how to decrease runtime for many classes~\cite{shen2021backdoor,xiang2022posttraining,hu2022trigger}, how many triggers can be recovered at once~\cite{hu2022trigger}, or how to reverse-engineer without computing gradients~\cite{dong2021black,guo2022aeva}.
 %Although many of these approaches rely directly on the networks gradients, 
 \citet{zhu2020gangsweep} establish that not an optimization, but also a GAN can be used to generate triggers. In general, a reconstruction can be based on the intuition that triggers themselves form distributions that can be learned~\cite{zhu2020gangsweep,qiao2019defending}.
Finally, \citet{liu2019abs} successfully use stimulation analysis of individual neurons to retrieve implanted trigger patterns.
Trigger reconstruction has been evaluated on almost all trigger-based backdoor attacks (see Table~\ref{table:matching_attack_defenses}), as their applicability is naturally limited to the existence of a trigger.

\subsection{Test Data Sanitization}\label{sec:testInputSanitization}
As the name suggests, this is the only group of defenses operating during \emph{test time}, where the defender attempts to sanitize malicious test inputs.
The assumptions on the defender's knowledge and capabilities, as in other cases, are diverse and encompass all possible settings. Test data sanitization can be used in all learning settings, where exceptions for specific cases might apply.
This group can, in principle, be applied in all learning scenarios, but is the only sanitization applicable if the model is only available as an online service, and accessible via queries.
There are three strategies overall when sanitizing test data. The first one boils down to removing the trigger~\cite{chou2018sentinet,doan2020februus,li2020rethinking}. For example \citet{chou2018sentinet} use interpretability techniques to identify crucial parts of the input and then mask these to identify whether they are adversarial or not. A second group is build on the agreement of ensembles on input~\cite{villarreal2020confoc,veldanda2020nnoculation,sarkar2020backdoor}. In \citet{sarkar2020backdoor}, this ensemble results indirectly from noising the input, but can also be build with a second, retrained version of the original model on different styles~\cite{villarreal2020confoc} or augmentations~\cite{veldanda2020nnoculation}. Finally, and as used for trigger generation, the consistency of a classifier's output can also help to detect an attack~\cite{gao2019strip,javaheripi2020cleann}. While \citet{gao2019strip} superimpose images to check the consistency, \citet{javaheripi2020cleann} instead consider the consistency of noised images in the inner layers.
As shown in Table~\ref{table:matching_attack_defenses}, test data sanitization has been tested only on trigger-based backdoor attacks. 
However, the latter two strategies mentioned above do have the potential to also detect targeted poisoning attacks, as these lead to locally implausible behavior. A detection of indiscriminate attacks at test time is however not possible.

\subsection{Current Limitations}
Although there is a large body of work on defenses, there are still unresolved challenges, as detailed in the following.

\subsubsection{Inconsistent Defense Settings}\label{sec::incDefenseSettings}
The assumptions on the defender's knowledge and capabilities reflect what is required to deploy a defense. In indiscriminate defenses, or robust training and training data sanitization in general, these are very homogeneous. When it comes to model inspection, trigger reconstruction, model sanitization, and test data sanitization, there is a larger variation in both the defender's knowledge and capabilities.
In particular, we lack understanding on the effect of individual capabilities or knowledge, for example not having direct access to the model when provided as a service and interacting via queries. More work is required that enables comparison across approaches here, and that sheds light on the individual components of the defense setting.

\subsubsection{Insufficient Defense Evaluations}\label{sec::DefEval} In Table~\ref{table:matching_attack_defenses}, we match poisoning attack strategies and defenses by reporting in each cell the defense papers that evaluate against the corresponding attack strategy. In some cases, indicated with a cross (\xmark\xspace), a defense of this strategy is not possible as there is no trigger to reconstruct (indiscriminate or targeted) or the test data is not altered by the attacker and can thus not be sanitized (indiscriminate attacks). Furthermore, 
Table~\ref{table:matching_attack_defenses} shows that the amount of defenses per attack strategies varies greatly.
Whereas for backdoor attacks using patch triggers there are around fifty defenses, only eleven defenses have been considered against semantic triggers, %~\cite{DBLP:conf/iccd/LiuXS17,shen2021backdoor,liu2018fine,zeng2020deepsweep,veldanda2020nnoculation}, 
  one against bilevel targeted attacks~\cite{frederickson2018attack}, one against bilevel patch backdoor attacks~\cite{yo2022CreatedEqual}, and none against indiscriminate clean label bilevel attacks.
With only a few defenses~\cite{zhu2021clear,tang2019demon}, there is also a shortage of model inspection and sanitization defenses when no trigger manifests in the model.\\
Beyond this shortage, there is a need to thoroughly test existing defenses using adaptive attacks, which are depicted in Table~\ref{tab:brokenDefs}.
Adaptive attacks are tailor to circumvent one or several defenses. In other words, the attack identifies essential components like for example a threshold within a defense and adapts the poisoning points to be below this threshold. For example, \citet{koh2018stronger} constrain the indiscriminate poisons features so that several points are in close vicinity to avoid outlier detection. In the case of backdoors, \citet{shokri2020bypassing} regularize the trigger to be less detectable within the network. \citet{tang2019demon} and \citet{junyuComposite} employ different strategies to make training data with trigger more similar to benign data. Yet, as visible in Table~\ref{tab:brokenDefs}, current adaptive backdoor attacks tend to break the same defenses. More work is thus needed to understand all defenses' limitations through  adaptive attacks, even though systematizing the design of such attacks and automating the corresponding evaluations is not trivial. To this end, it may be interesting to design \textit{indicators of failure} that automate the identification of faulty, non-adaptive evaluations for poisoning attacks, as recently shown in~\citep{pintor2021indicators} for adversarial examples.
 
 \begin{table}[t]
    \centering
    \caption{Attacks breaking defenses in the areas of indiscriminate, targeted, and backdoor attacks. We provide the reference for the adaptive attack, which defenses are broken, and a high-level description of the strategy of the adaptive attack.}\label{tab:brokenDefs}
    \begin{tabular}{lllll}
        \toprule
%        & &  \multicolumn{3}{r}{Defender's knowledge}& \multicolumn{4}{l}{and capabilities} \\
%        \cmidrule(r){4-5}  \cmidrule(r){6-7}  
& \multicolumn{3}{c}{Broken Defenses} \\
\cmidrule(r){2-4}
 Attack & Indiscriminate & Targeted & Backdoor & Strategy\\ 
 \midrule
 \citet{koh2018stronger} &  \cite{2017arXiv170603691S,rubinstein2009antidote} &&& constrain poison point's features \\
 \citet{shokri2020bypassing} & & & \cite{wangneural,tran2018spectral,chen2018detecting} & regularize trigger pattern\\
 \citet{tang2019demon} & &  & \cite{wangneural,chou2018sentinet,gao2019strip,chen2018detecting} & add trigger images with correct label \\
 \citet{junyuComposite} & & & \cite{wangneural,liu2019abs} & add trigger images mixed from source and target \\
 \bottomrule
 \end{tabular}
 \end{table}
 \subsubsection{Overly-specialized Defenses.}\label{sec:specificDefenses} 
 Furthermore, few defenses (only roughly one sixth) have been evaluated against different kinds of triggers.
 Only one defense in test data sanitization~\cite{zeng2020deepsweep} and two defenses in trigger reconstruction~\cite{liu2019abs,hu2022trigger} have been evaluated against more than one trigger type.
 There are three defenses for each training data sanitization~\cite{xiang2021reverse,shawn2022traceback,hayase2021spectre}, model sanitization~\cite{li2021neural,zeng2020deepsweep,zeng2021adversarial} and robust training~\cite{borgnia2021dp,geiping2021doesn,li2021anti}. In model inspection,  five~\cite{soremekun2020exposing,tang2019demon,Xu2021MetaAudio,shen2021backdoor,guo2022aeva} defenses tests on more than one attack type.
 There are even more general  defenses that are able to handle multiple poisoning attacks, such as indiscriminate, targeted, and backdoor attacks, as for example \citet{geiping2021doesn} and \citet{hong2020effectiveness} show.
 
\section{Poisoning Attacks and Defenses in Other Domains}\label{sec:other_domains}
While in this survey we focus on poisoning ML in the context of supervised learning tasks, and mostly in computer-vision applications, it is also worth remarking that several poisoning attacks and defense mechanisms have been developed also in the area of federated learning~\cite{Tolpegin20federflip, Cao19FederFlip,Zhang19GanFeder,Bagdasaryan20BackdoorFeder,Bhagoji19BackdoorFeder,Sun19BackdoorFeder,Xie20DBA,zhang2017game,zhang2020defending,hayes2018contamination,zhao2019pdgan}, regression learning~\cite{feng2014regression,Liu2017RobustLR,jagielski2018manipulating,muller2020regression,diakonikolas2019sever,wen2021great}, reinforcement learning~\cite{kiourti2020trojdrl,behzadan2017vulnerability,zhang2020adaptive,huang2019deceptive,rakhsha2020policy,wang2020reinforcement,xu2021transferable,ashcraft2021poisoning,ma2018data,banihashem2021defense,everitt2017reinforcement,han2020adversarial}, and unsupervised clustering~\cite{rubinstein2009antidote,kloft2010anomaly,Biggio2013IsDC,Biggio2014PoisoningBM,cina2022clustering} or anomaly detection~\cite{rubinstein2009antidote,Cretu2008Casting-out-Dem} algorithms.
Furthermore, notable examples of poisoning attacks and defenses have also been shown in computer-security applications dealing with ML, including spam filtering~\cite{nelson08spam,biggio2011bagging,diakonikolas2019sever,paudice2018label,frederickson2018attack}, network traffic analysis~\cite{rubinstein2009antidote}, and malware detection~\cite{perdisciWormSignature2006,severi_explanation-guided_2021,taheri2020defending}, audio~\cite{venomave_aghakhani_2020,liu_trojaning_2018,Xu2021MetaAudio,Koffas2021Ultrasonic,liu2018fine} and video analysis~\cite{zhao_clean_video_2020,tran2018spectral}, natural language processing~\cite{liu_trojaning_2018,badnl_chen_2020,trojaning_nlp_zhang_2020,qi2020onion,chen2020mitigating}, and even in graph-based ML applications~\cite{Liu19GraphSSL, Zugner19GraphSSL,BojchevskiG19NodeEmbedding,Zhang21FixedBackdoorGraph,Xi21GraphBackdoor}. While, for the sake of space, we do not give a more detailed description of such research findings in this survey, we do believe that the systematization offered in our work provides a useful starting point for the interested reader to gain a better understanding of the main contributions reported in these other research areas.

\section{Resources: Software Libraries, Implementations, and Benchmarks}
\label{sec:resources}

Unified test frameworks play a huge role when evaluating and benchmarking both poisoning attacks and defenses. We thus attempt to give an overview of available resources in this section.
Libraries and available code ease the evaluation and benchmarking of both attacks and defenses. Ignoring the many repositories containing individual attacks, to date, only a few libraries provide implementations of poisoning attacks and defenses.\footnote{Analysis carried out in June 2022.}
The library with the largest number of attacks and defenses is the adversarial robustness toolbox (ART)~\cite{art2018}. ART implements indiscriminate poisoning attacks~\cite{biggio2012poisoning}, targeted~\cite{shafahi_poison_2018,aghakhani20bullseye,turner_clean-label,geiping_witches_2020} and backdoor attacks~\cite{gu_badnets_2017,saha_hidden_2019}, as well as an adaptive backdoor attack~\cite{shokri2020bypassing}. The library further provides a range of defenses~\cite{chen2018detecting,baracaldo2018detecting,nelson2009misleading,tran2018spectral,wangneural,gao2019strip}.
Furthermore, SecML~\cite{melis2019secml} provides indiscriminate poisoning attacks against SVM, logistic, and ridge regression~\cite{biggio2012poisoning,demontis2019whytransferability,battista_secure_features2015}. Finally, the library advBox~\cite{goodman2020advbox} provides both indiscriminate and backdoor attacks on a toy problem. 

Beyond the typical ML datasets that can be used for evaluation, there exists a large database from the NIST competition,\footnote{\url{https://pages.nist.gov/trojai/docs/index.html}} %. The database 
which contains a large number of models from image classification, object recognition, and reinforcement learning. Each model is labeled as poisoned or not. 
The module further allows to generate new datasets with poisoned and unpoisoned models. %Last but not least, 
\citet{schwarzschild2021just} recently introduced a framework to compare different poisoning attacks. They conclude that for many attacks, the success depends highly on the experimental setting.
%
%\myparagraph{Conclusion.} 
To conclude, albeit a huge number of attacks and defenses have been introduced, there is still a need of libraries that allow access to  off-the-shelf implementations to compare new approaches. In general, few works benchmark poisoning attacks and defenses or provide guidelines to evaluate  poisoning attacks or defenses.  
%\end{comment}

\section{Development, Challenges, and Future Research Directions}
\label{sec:development}

In this section, we outline challenges and future research directions for poisoning attacks and defenses. We start by discussing the intertwined historical development of attacks and defenses, and then highlight the corresponding challenges, open questions, and promising avenues for further research.
\begin{figure}[t]
\centering
\includegraphics[width=1\textwidth]{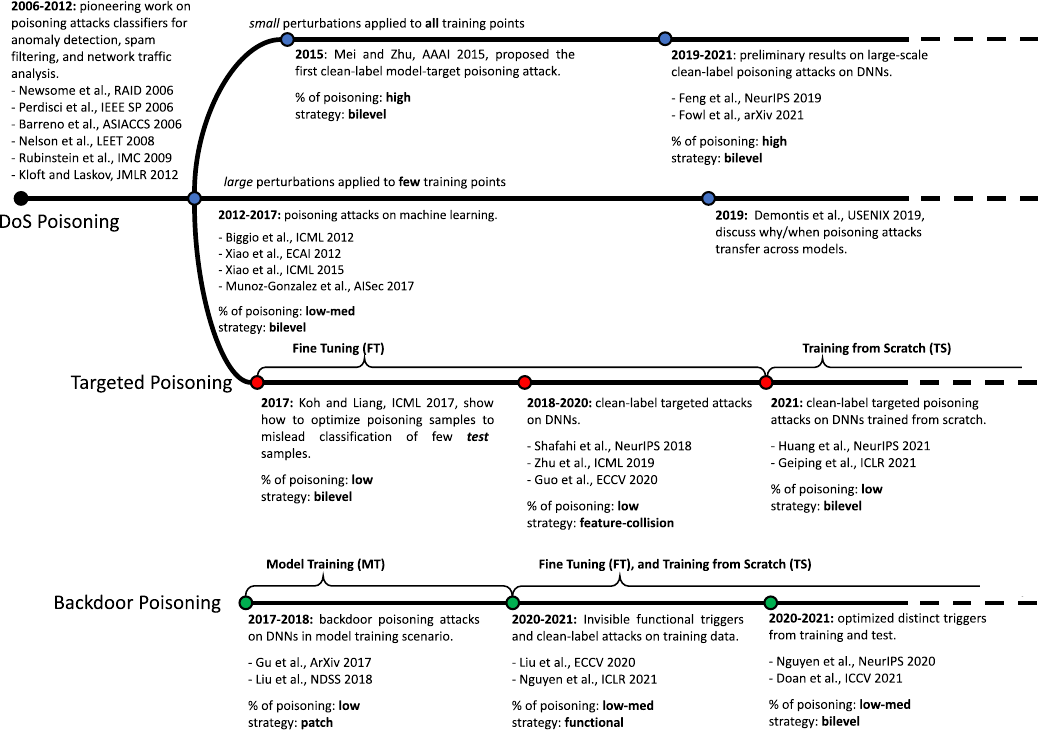}
\caption{Timeline for indiscriminate (blue), targeted (red) and backdoor (green) data poisoning attacks on machine learning. Related work is highlighted with markers of the same color and connected with dashed lines to highlight independent (but related) findings. %The date of publication refers to publication in peer-reviewed conferences and journals, \edit{or to the pre-print publication, if the article has not yet been published}.
}
\label{fig:timeline_attacks}
\end{figure}

\subsection{Development Timelines for Poisoning Attacks and Defenses}
\label{subsect:devel}

\begin{figure}[t]
\centering
\includegraphics[width=1\textwidth]{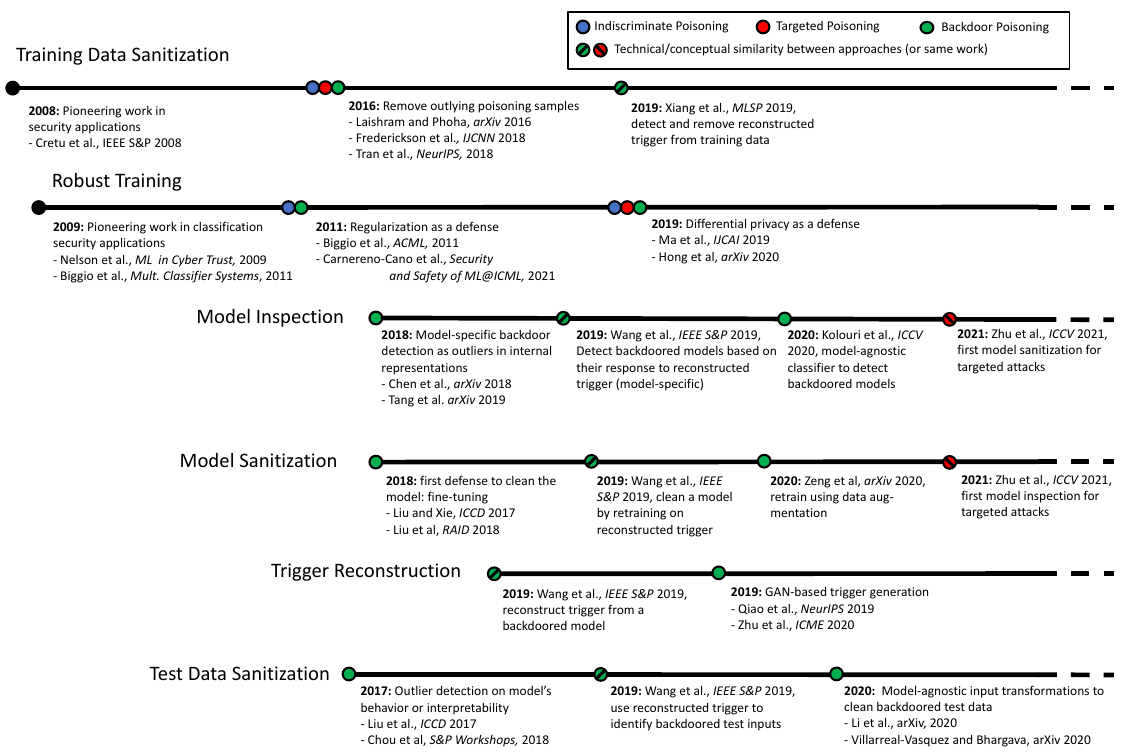}
\caption{Timeline of the six kinds of defenses described in Sect.~\ref{sec::defenses}. The dots remark against which class of attacks defenses have been introduced, dashed lines denote related approaches, and thin gray lines connect the same work across defense groups.}
\label{fig:timeline_defenses}
\end{figure}

We start by discussing the historical development of poisoning attacks (represented in Fig.~\ref{fig:timeline_attacks}), and afterwards that of defenses (depicted in Fig.~\ref{fig:timeline_defenses}). In both cases, we highlight the respective milestones and development over time.

\subsubsection{Attack Timeline.}The attack timeline is shown in Fig.~\ref{fig:timeline_attacks}.
To the best of our knowledge, the first example of indiscriminate poisoning was developed in 2006 by \citet{perdisciWormSignature2006}, \citet{barreno2006anomaly}, and \citet{NewsomeParagraph2006}  in the computer security area. Such attacks, as well as subsequent attacks in the same area \cite{rubinstein2009antidote,kloft2010anomaly}, were based on heuristic approaches to mislead application specific ML models, and there was not a unifying mathematical formulation describing them. %to mislead a worm signature generator, i.e., in the context of a specific security application in which machine learning was used to generate signatures to detect computer worm viruses. 
It was only later, in 2012, that indiscriminate poisoning against machine learning was formulated for the first time as a bilevel optimization~\cite{labelflip_xiao_2012}, to compute optimal label-flip poisoning attacks. Since then, indiscriminate poisoning has been studied under two distinct settings, i.e., assuming either (i) that a small fraction of training samples can be largely perturbed~\citep{biggio2012poisoning,munoz-gonzalez_towards_2017,demontis2019whytransferability}; or (ii) that all training points can be slightly perturbed~\citep{using_machine_learning,confuse_fengCZ_2019,preventing_fowl_2021}. 

Targeted and backdoor poisoning attacks only appeared in 2017, and interestingly, they both started from different strategies. Targeted poisoning started with the bilevel formulation in~\citet{koh_understanding_2017}, but evolved in more heuristic approaches, such as feature collision~\cite{shafahi_poison_2018,zhu_transferable_2019,guo_practical_2020}. Only recently, targeted poisoning attacks were reformulated as bilevel problems, given the limitation of the aforementioned heuristic approaches~\cite{huang_metapoison_2020,geiping_witches_2020}. Backdoor poisoning started with the adoption of patch~\cite{gu_badnets_2017,liu_trojaning_2018} and functional~\cite{reflection_backdoor_liu_2020,input_aware_nguyen_2020} triggers. However, in the last years, such heuristic choices have been put aside, and backdoor attacks are getting closer and closer to the idea of formulating them in terms of a bilevel optimization, not only to enhance their effectiveness, but also their ability to bypass detection~\cite{saha_hidden_2019,sleeper_goldstein_2021}.

The historical development of the three types of attacks is primarily aimed at solving or mitigating as much as possible the challenges highlighted in Sect.~\ref{sec:attack_challenges}, i.e., (i) considering more realistic threat models, and (ii) designing more effective and scalable poisoning attacks.
In particular, recent developments in attacks seek to improve the applicability of their threat models, by tampering with the training data as little as possible (e.g., a few points altered with invisible perturbations) to evade defenses, and by considering more practical settings (e.g, \emph{training-from-scratch}). Moreover, more recent poisoning attacks aim to tackle the computational complexity and time required to solve the bilevel problem, not only to improve attack scalability but also their ability to stay undetected against current defenses. In Sect.~\ref{sec:challenges_future_developments} we more thoroughly discuss these challenges, along with some possible future research directions to address them.

\subsubsection{Defense Timeline.} The defense timeline is shown in Fig.~\ref{fig:timeline_defenses}.
%historic start
The first defenses, training data sanitization and robust training variants, were introduced %2009 
2008 and 2009 in a security context~\cite{nelson2009misleading,biggio11labelnoise,Cretu2008Casting-out-Dem}.
%training data sanitization without trigger removal
Following works in training data sanitization were based on outlier detection, and mitigated backdoor~\cite{tran2018spectral}, indiscriminate~\cite{DBLP:journals/corr/LaishramP16} and targeted~\cite{frederickson2018attack} attacks.
% all robust training
To train robustly, \citet{biggio2011bagging} showed 2011 that regularization can serve as a defense, a finding recently confirmed for backdoors~\cite{carnerero2021regularization}. In 2019,  differential privacy was shown to be able to mitigate  poisoning attacks~\cite{private_learners_ma_2019,hong2020effectiveness}. %, at the cost of reduced accuracy. 
This connection to privacy underlines the need to study poisoning also in relation to other ML security issues, as we will discuss in Sect.~\ref{sec::multAttacks}.
%start talking about the remaining backdoor based defenses
The remaining kinds of defenses are characterized by more diverse threat models, as we discussed in
Sect.~\ref{sec::incDefenseSettings}. The type of attack mitigated is however less diverse, and focuses mainly on backdoors, as explained in Sect.~\ref{sec:specificDefenses}.
%model inspection
We start with model inspection approaches, which were first introduced by \citet{chen2018detecting} and were based on outlier detection on latent representations. In 2020, \citet{kolouri2020universal} generalized the backdoor inspections to be model independent using a meta-classifier. Recently, \citet{zhu2021clear} introduced a search-based approach to determine whether a model suffers from targeted poisoning.
%use Zhu et al to make connection to model sanitization
The latter approach also proposed how to sanitize the model. 
%model sanitization
The first defenses for such model sanitization against backdoors were trigger agnostic and based on fine-tuning~\cite{DBLP:conf/iccd/LiuXS17,liu2018fine} and later on data augmentation~\citet{zeng2020deepsweep}. 
 %trigger reconstruction
 Another possibility, introduced by \citet{wangneural}, is to retrain a model based on a reconstructed trigger. 
 \citet{wangneural} introduced the idea of reconstruction a trigger in 2019. They generated a trigger based on optimization of a pattern that causes backdoor behavior, e.g., misclassification of many samples when added to them. More recent approaches improve trigger reconstruction by considering distributions over triggers~\cite{qiao2019defending}, not individual patterns. 
A reconstructed trigger can also serve to inspect a model~\cite{wangneural},
%trigger reconstruction + relation to remaining defenses
%test data sanitization
or serve to sanitize test data ~\cite{wangneural}. However, the first approaches to sanitize test data in 2017 were based on outlier detection to inspect the model inspection and sanitize training data. Analogous to model inspection, initial works relied on latent, model specific features~\cite{DBLP:conf/iccd/LiuXS17} whereas later works from 2020 use model-agnostic input transformations~\cite{li2020rethinking}.

%The first training data sanitizations (based on outlier detection) are an exception, and mitigate not only backdoor~\cite{tran2018spectral}, but also indiscriminate ~\cite{DBLP:journals/corr/LaishramP16} and  targeted attacks~\cite{frederickson2018attack}. %The latter work is of particular importance, as it was the first to point out a possible trade-off between attacks strength and detection, more in detail discussed in Sect.~\ref{sec:genChallenges}. 
%An additional possibility to sanitize the training data from backdoors is to remove a known trigger~\cite{xiao2019characterizing}.
%This leads to the group of defenses enabling trigger reconstruction. In 2019, \citet{wangneural} were the first to show that a trigger can be reconstructed. More recent approaches improve by considering distributions over triggers~\cite{qiao2019defending}, not individual patterns. 
%A reconstructed trigger can also serve as proof that a model is backdoored~\cite{wangneural}. 

%We can also attempt to remove incoming triggers in test samples, or sanitize the test data. First test data  sanitizations~\cite{DBLP:conf/iccd/LiuXS17} from 2017 were, like for model inspection and training data sanitization,  based on outlier detection. Analogous to model inspection, initial works relied on latent, model specific features~\cite{DBLP:conf/iccd/LiuXS17} whereas later works from 2020 use model-agnostic input transformations~\cite{li2020rethinking}. Also analogous to for example model inspection, a reconstructed trigger can easily be removed from incoming test data~\cite{wangneural}.
 %adaptive attacks
One historical development which is highly relevant but left out in both timeline figures is the study of adaptive attacks against defenses to assess their robustness, as discussed in Sect.~\ref{sec::DefEval}. We elaborate on this challenge in Sect.~\ref{sec::adaptiveAttacks}.

\subsection{Challenges and Future Work}~\label{sec:challenges_future_developments}
Building on the development timelines and the corresponding overview provided in Sect.~\ref{subsect:devel}, we formulate some future research challenges for both poisoning attacks and defenses in the remainder of this section.

\subsubsection{Considering Realistic Threat Models.} 
%%% BEFORE REBUTTAL
%One pertinent challenge arising from the discussion on poisoning attacks in Sect.~\ref{sec:attack_challenge_unrealistic} demands considering more realistic threat models and attack scenarios, as also recently pointed out in~\cite{shej22-sp}. While assessing machine learning models in real-world settings is not straightforward~\cite{sommer2010outside}, the need to develop realistic threat models, possibly for individual applications, is still an open question in machine learning security, and has so far only received recognition for test-time attacks~\cite{gilmer2018motivating}.
%We would thus invite the research community to evaluate poisoning attacks under more realistic assumptions, which also take into account the specific application domain. 
%%% After REBUTTAL
One pertinent challenge arising from the discussion on poisoning attacks in Sect.~\ref{sec:attack_challenge_unrealistic} demands considering more realistic threat models and attack scenarios, as also recently pointed out in~\cite{shej22-sp}. While assessing machine learning models in real-world settings is not straightforward~\cite{sommer2010outside}, the need to develop realistic threat models is still an open question in machine learning security and has so far only received recognition for test-time attacks~\cite{gilmer2018motivating}.
Here we define some guidelines that can serve as a basis for future work that wants to assess the real safety impact of poisoning versus real applications. First, limit the attacker's knowledge of the target system and their capacity to tamper during training. For example, an attack that assumes only a small percentage of control over the training set can be broadly applied. Second, develop more stealthy poisoning strategies to avoid detection against defenses. Some attack strategies, \eg, patch trigger or feature collision, are computationally efficient, but several defensive countermeasures exist to detect them (see Table~\ref{table:matching_attack_defenses}). Finally, evaluating poisoning attacks against real-world applications and making them adaptive to the presence of a defender. 
Therefore, we invite the research community to evaluate poisoning attacks with more realistic or less favorable assumptions for the attacker, which also take into account the specific application domain.

\subsubsection{Designing More Effective and Scalable Poisoning Attacks}
%The other challenge we highlighted in Sect.~\ref{sec:attack_challenge_scalability} is the computational complexity of poisoning attacks when relying on bilevel optimization. However, the same limitation is also encountered in other research domains such as hyperparameter optimization and meta-learning. More concretely, the former is the process of determining the optimal combination of hyperparameters that maximizes the performance of an underlying learning algorithm. On the other hand, meta-learning encompasses feature selection, algorithm selection, learning to learn, or ensemble learning, to which the same reasoning applies. 
The other challenge we highlighted in Sect.~\ref{sec:attack_challenge_scalability} is the computational complexity of poisoning attacks when relying on bilevel optimization. However, the same limitation is also encountered in other research domains such as hyperparameter optimization and meta-learning which naturally are formulated within the mathematical framework of bilevel programming~\cite{franceschi18BilevelHM}. More concretely, the former is the process of determining the optimal combination of hyperparameters that maximizes the performance of an underlying learning algorithm. On the other hand, meta-learning encompasses feature selection, algorithm selection, learning to learn, or ensemble learning, to which the same reasoning applies. 
Having formulated poisoning attacks within the bilevel framework (see Sect.~\ref{sec:unifying_framework}) hints that strategies developed to speed up the optimization of bilevel programs involved in meta-learning or hyperparameter optimization taks can be adapted to facilitate the development of novel scalable attacks.
In principle, by imagining poisoning samples as the attacker-controlled learning hyperparameters, we could apply the approaches proposed in these two fields to mount an attack. Notably, we find some initial works connecting these two fields with data poisoning.
For example, \citet{shen2021backdoor} rely in their approach on a k-arms technique, a technique similar to bandits, as done by~\citet{jones1998efficient}. 
Further, \citet{munoz-gonzalez_towards_2017} exploited the back-gradient optimization technique proposed in \cite{Maclaurin15Backgradient,Domke12Hypergrad}, originally proposed for hyperparameter optimization, and subsequently, \citet{huang_metapoison_2020} inherited the same approach making the attack more effective against deep neural networks.
Apart from the work just mentioned, the connection between the two fields and poisoning is still currently under-investigated, and other ideas could still be explored. For example, the optimization proposed by \cite{lorraine2020optimizing} can further reduce run-time complexity and memory usage even when dealing with millions of hyperparameters. Or another way might be to move away from gradient-based approaches and consider gradient-free approaches, thus overcoming the complexity of the inverting the Hessian matrix seen in Sect.~\ref{sec:attack_challenge_scalability}. In the area of gradient-free methods, the most straightforward way is to use grid or random search~\cite{bergstra2012random}, which can be sped up using reinforcement learning~\cite{li2017hyperband}. 
Also, Bayesian optimization has been used, given a few sampled points from the objective and constraint functions, to approximate the target function~\cite{jones1998efficient}. Last but not least, evolutionary algorithms~\cite{young2015optimizing} as well as particle swarm optimization~\cite{lorenzo2017particle} have shown to be successful.

In conclusion, we consider these two fields as possible future research directions to find more effective and scalable poisoning attacks for assessing ML robustness in practice.% and practical test machine learners' robustness. 

\subsubsection{Systematizing and Improving Defense Evaluations.}\label{sec::adaptiveAttacks} 
Regardless of future attacks, we need to systematize and understand the limits of existing (and future) defenses better.
As we have seen in Sect.~\ref{sec:resources}, there is no coherent benchmark for defenses. Such a benchmark exposes flawed evaluations and assesses the robustness of a defense per se or in relation to other defenses (taking into account the defense's setting, as discussed in Sect.~\ref{sec::DefEval}). Jointly with  benchmarks, evaluation guidelines, as discussed for ML evasion by \citet{carlini2019evaluating}, help to improve defense evaluation.
More specifically, these guidelines can encompass knowledge when attacks fail and why, similar to work on evasion attack failure~\cite{pintor2021indicators}. 
Crucial in this context is also, as discussed in Sect.~\ref{sec::DefEval}, to expand our understanding of adaptive attacks.

An orthogonal question is how to increase existing knowledge about trade-offs
between for example attack strength and stealthiness for indiscriminate attacks~\cite{frederickson2018attack} or backdoors~\cite{turner_clean-label, chen_targeted_2017, dynamic_backdoor_salem_2020}. 
Further trade-offs relate clean accuracy and accuracy under the poisoning attack by hyperparameter tuning. More concretely, \citet{demontis2019whytransferability} and \citet{cina2021backdoor} showed that more regularized classifiers tend to resist better to poisoning attacks, at the cost of slightly reducing clean accuracy. 
Ideally, impossibility results further increase our knowledge about hard limitations. 
To the best of our knowledge, the only impossibility results provided thus far for subpopulation poisoning attacks can be found in~\cite{subpopulation_jagielski_2020}, showing that it is impossible to defend poisoning attacks that target only a fraction of the data. Expanding our knowledge about trade-offs and impossibilities will help to design and configure effective defenses.

\subsubsection{Designing Generic Defenses against Multiple Attacks.}\label{sec::multAttacks} 
Such effective defenses also need to overcome, as discussed in the Sect.~\ref{sec:specificDefenses}, that they often specialize and to on one or several poisoning attacks (for example backdoor and targeted). Such one-sided evaluations introduce a bias, and the effect of this overfitting on biased datasets has been recognized in image recognition~\cite{torralba2011unbiased}, but received relatively little attention in security so far. Some defenses, however, do evaluate several poisoning attacks~\cite{shen2021backdoor,geiping2021doesn,hong2020effectiveness}, or even different ML security threats like backdoors and evasion~\cite{sun2021can} or poisoning and privacy~\cite{hong2020effectiveness}.

In addition to creating more robust defenses, such interdisciplinary works also increase our understanding of how poisoning interferes with non-poisoning ML attacks~\cite{weng2020trade,chase2021property}.
One attack is evasion, where a small perturbation is added to a sample at test time to force the model to misclassify an output.
 Evasion is closely related, but different from backdoors, which add a \emph{fixed} perturbation at \emph{training time}, causing an upfront known vulnerability at test time. 
Only a few works study evasion and poisoning together. 
For example, \citet{sun2021can} introduce a defense against both backdoors and adversarial examples.
Furthermore, \citet{fowl2021adversarial} show that adversarial examples with the original labels are strong poisons at training time.
In the opposite direction, \citet{weng2020trade} %study the relationship between evasion and clean label backdoor vulnerability. They 
find that if backdoor accuracy is high, evasion tends to be less successful and vice versa.
Furthermore, \citet{mehra2021robust} study poisoning of certified evasion defenses: using poisoning, they decrease the certified radius and accuracy.
Two works, namely by \citet{manoj2021excess} and \citet{goldwasser2022planting}, relate evasion and backdoors in a theoretical way. Both share rigid assumptions, however \citet{manoj2021excess} show an impossibility results in terms of non-existence of backdoor for some natural learning problems. \citet{goldwasser2022planting}, on the other hand, show that backdoor detection might be impossible.
In relation to privacy or intellectual property, poisoning can be used to increase the information leakage from training data at test time on collaborative learning~\cite{chase2021property}.
%Poisoning has also recently been used to circumvent fairness~\cite{solans2020poisoning-fairness} and studied in relation to privacy. 
Privacy can further be a defense against poisoning~\cite{hong2020effectiveness, borgnia2021dp}, or poisoning can be a tool to obtain~\cite{preventing_fowl_2021}. % or to circumvent privacy~\cite{private_learners_ma_2019}.
Summarizing, there is little knowledge on how poisoning  interacts with other attacks. More work is needed to understand this relationship and secure machine learning models in practice against several threats at the same time.

\section{Concluding Remarks}\label{sec::conclusion}
\begin{comment}
In this survey, we reviewed the state of the art in poisoning attacks and defenses. To conclude, %this survey,
we now briefly review each of the three areas with their corresponding state of the art and challenges. In the area of indiscriminate poisoning, the understanding of both attacks and defenses for convex learners is rather complete. Open questions concern here the feasibility of attacks against deep learning, as well as whether defenses at test time are possible.
In the area of targeted poisoning, the challenge to poison DNN is partially solved, with the limitation that feature collision attacks are still not robust to changes in the optimizer, data augmentation, and do not work when training from scratch. In this area, more research on defenses is required. Finally, in backdoor attacks, plenty of  heuristic attacks and  defenses have been proposed. Stealthier triggers require more poisoning samples, and are harder to protect against.
Although a large body of defenses exists, there is an ongoing arms race between attackers and defenders, and current limitations of defenses, in terms of capabilities and threat modeling, are not well understood. Furthermore,  most poisoning attacks are mounted in feature space, and our knowledge on their impact in the physical world is currently limited.
Finally, more work is  needed to design robust defenses against multiple poisoning attacks, to protect ML against poisoning and other security threats, and to obtain certifiable security guarantees. 
\end{comment}

The increasing adoption of data-driven models in production systems demands a rigorous analysis of their reliability in the presence of malicious users aiming to compromise them. Within this survey, we systematize a broad spectrum of data poisoning attacks and defenses according to our modeling framework, and we exploit such categorization to match defenses with the corresponding attacks they prevent. Moreover, we provide a unified formalization for poisoning attacks via bilevel programming, and we spotlight resources (e.g., software libraries, datasets) that may be exploited to benchmark attacks and defenses. Finally, we trace the historical development of data literature since the early developments dating back to more than 20 years ago and find the open challenges and possible research directions that can pave the way for future development. 
In conclusion, we believe our contribution can help clarify what threats an ML system may encounter in adversarial settings and encourage further research developments in deploying trustworthy systems even in the presence of data poisoning threats.
\begin{acks}
This work has been partly supported by: the PRIN 2017 project RexLearn (grant no. 2017TWNMH2), funded by the Italian Ministry of Education, University and Research; the EU's Horizon Europe research and innovation program under the project ELSA, grant agreement No 101070617; the project ``TrustML: Towards Machine Learning that Humans Can Trust’’, CUP: F73C22001320007, funded by Fondazione di Sardegna; the NRRP MUR program funded by the EU - NGEU under the project SERICS (PE00000014); and the COMET Programme managed by FFG in the COMET Module S3AI, funded by BMK, BMDW, and the Province of Upper Austria.
\end{acks}

\bibliographystyle{ACM-Reference-Format}
%\bibliography{lit, bibDB}
\bibliography{lit}

%\appendix 
%\input{sections/poisoning_other_domains}
\end{document}